\documentclass[a4paper,fleqn, onecolumn]{cas-sc}

\usepackage[authoryear,longnamesfirst]{natbib}

\usepackage{tabularx}
\usepackage[flushleft]{threeparttable}
 \usepackage[ruled,linesnumbered]{algorithm2e}
\usepackage{graphicx}
\definecolor{codered}{rgb}{0.98,0.49,0.72}
\def\tsc#1{\csdef{#1}{\textsc{\lowercase{#1}}\xspace}}
\tsc{WGM}
\tsc{QE}
\tsc{EP}
\tsc{PMS}
\tsc{BEC}
\tsc{DE}

\begin{document}
\let\WriteBookmarks\relax
\def\floatpagepagefraction{1}
\def\textpagefraction{.001}
\shorttitle{}
\shortauthors{Wei Luo et~al.}

\title [mode = title]{Template-Based Feature Aggregation Network for Industrial Anomaly Detection}                      



\author[1]{Wei Luo}[orcid=0000-0003-3125-054X]
\ead{luow23@mails.tsinghua.edu.cn}


\address[1]{State Key Laboratory of Precision Measurement Technology and Instruments, Tsinghua University, Beijing, 100084, China}

\author[1]{Haiming Yao}[orcid=0000-0003-1419-5489]
\ead{yhm22@mails.tsinghua.edu.cn}

\author[2]{Wenyong Yu}[
    orcid=0000-0003-4012-1264
   ]
\cormark[1]
\ead{ywy@hust.edu.cn}


\address[2]{State Key Laboratory of Digital Manufacturing Equipment and Technology, Huazhong University of Science and Technology, Wuhan, 430074, China}

\cortext[cor1]{Corresponding author}


\begin{abstract}
Industrial anomaly detection plays a crucial role in ensuring product quality control. Therefore, proposing an effective anomaly detection model is of great significance. While existing feature-reconstruction methods have demonstrated excellent performance, they face challenges with shortcut learning, which can lead to undesirable reconstruction of anomalous features. To address this concern, we present a novel feature-reconstruction model called the \textbf{T}emplate-based \textbf{F}eature \textbf{A}ggregation \textbf{Net}work (TFA-Net) for anomaly detection via template-based feature aggregation. Specifically, TFA-Net first extracts multiple hierarchical features from a pre-trained convolutional neural network for a fixed template image and an input image. Instead of directly reconstructing input features, TFA-Net aggregates them onto the template features, effectively filtering out anomalous features that exhibit low similarity to normal template features. Next, TFA-Net utilizes the template features that have already fused normal features in the input features to refine feature details and obtain the reconstructed feature map. Finally, the defective regions can be located by comparing the differences between the input and reconstructed features. Additionally, a random masking strategy for input features is employed to enhance the overall inspection performance of the model. Our template-based feature aggregation schema yields a nontrivial and meaningful feature reconstruction task. The simple, yet efficient, TFA-Net exhibits state-of-the-art detection performance on various real-world industrial datasets. Additionally, it fulfills the real-time demands of industrial scenarios, rendering it highly suitable for practical applications in the industry. Code is available at \href{https://github.com/luow23/TFA-Net}{\textcolor{codered}{https://github.com/luow23/TFA-Net}}
\end{abstract}




\maketitle

\section{Introduction}
\begin{figure}[!t]
    \centering
    \includegraphics{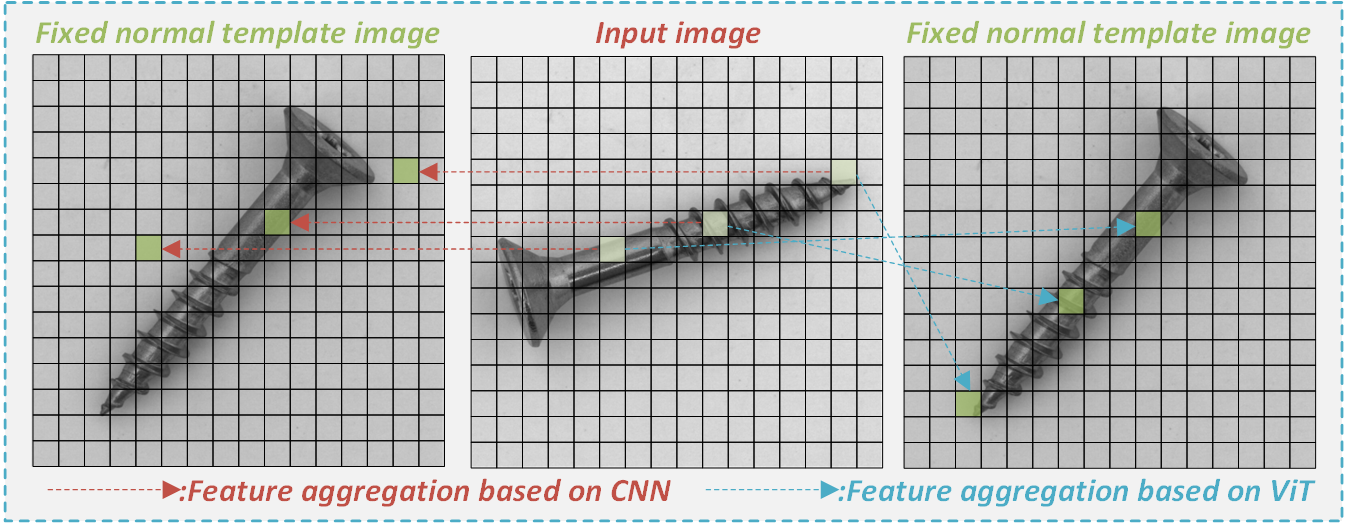}
    \caption{The comparison of feature aggregation methods based on CNN and ViT. {For images from various orientations, CNN models may face challenges in feature aggregation due to their translational equivariance induction bias. In contrast, ViT exhibits better feature aggregation capabilities as it lacks the inductive bias of CNN and possesses global modeling capabilities.}}
    \label{fig:VIT-CNN-COM}
\end{figure}

Visual anomaly detection (VAD) is a critical process that involves the detection of anomalies \cite{yang2023memseg, cai2023itran, fan2023transferable} which significantly deviate from normal data. {It plays a crucial role in industrial product quality control \cite{industrial1, industrial2} and ensuring an exceptional user experience, thereby making it an indispensable aspect of practical industrial production.}\\
\indent Although the field of VAD has recently gained considerable attention, achieving accurate anomaly detection remains a challenging task. The diversity of defects in terms of size and category makes it difficult for traditional methods \cite{TEXTEMS,LCA,PHOT} to achieve adaptive optimization. Furthermore, due to the unpredictable nature of defects, collecting and annotating a dataset that encompasses all types of defects is impractical, thereby limiting the application of supervised methods \cite{PGANet, RetinaNet}. Therefore, VAD task is typically performed in an unsupervised manner.\\
\indent Unsupervised anomaly detection methods can be broadly categorized into two groups: embedding and reconstruction-based approaches. Embedding-based approaches \cite{PatchSVDD, SPADE, MBPFM} transform the images into a discriminative embedding space where the embedding distance between normal and anomaly is large. However, the high cost in terms of memory requirement and low inference speed of embedding-based methods limit their real-world industrial applications. In contrast, reconstruction-based methods show great potential in practical industrial applications. As a classical reconstruction model, the autoencoder (AE) \cite{AE} compresses images into latent features and utilizes them to reconstruct the images. Given that only normal samples are employed during training, a larger reconstruction error is anticipated in the defective regions while a smaller reconstruction error is expected in the normal regions during testing. Nonetheless, the high generalization ability of vanilla AE may lead to perfect reconstruction of defects, resulting in low inspection accuracy. To mitigate this issue, MemAE \cite{MemAE} utilizes a memory bank to store typical normal prototypes during training. During testing, the model retrieves the normal prototype from the memory bank that is most similar to the tested features in order to obtain defect-free results. Numerous studies \cite{DAAD,TrustMAE,sa-memory,pmemory, CMA-AE} have been conducted to improve memory mechanisms, yet their efficacy has been inconclusive. Recently, generative adversarial networks (GANs) \cite{GANs} have gained wide popularity in anomaly detection due to their powerful generation capabilities. As an example, Yang \textit{et al}. \cite{AFEAN} proposed a GAN-based framework that edits anomalous features to reconstruct normal backgrounds. In another work, Yao \textit{et al}. \cite{FMR-Net} presented a feature memory rearrangement network that leverages memory-generated features to repair anomalous features. {CDO \cite{CaoTii} employs a mask strategy to increase the margin between normal and anomaly distributions, thereby enhancing the model's detection performance. RIAD \cite{RIAD} introduces the use of masks of varying sizes to transform the reconstruction task into the image inpainting task.}\\
\indent However,
\begin{figure}[!t]
    \centering
    \includegraphics[width=83mm]{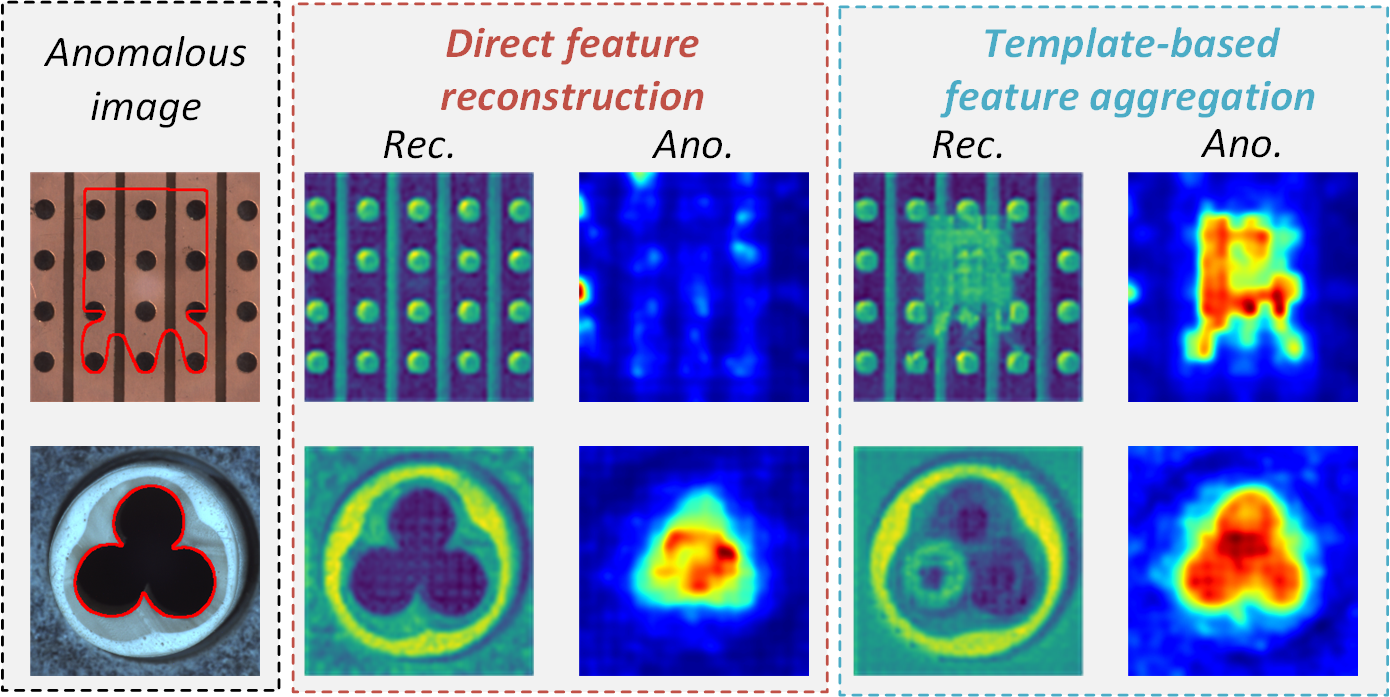}
    \caption{The comparison between direct feature reconstruction and template-based feature aggregation. Rec. and Ano. denote the reconstructed feature and anomaly map, respectively.}
    \label{fig:global-defect}
\end{figure}the aforementioned reconstruction-based methods are still fall short in terms of detection performance, as accurately reconstructing the background is challenging and the pixel differences between input and reconstructed images do not carry semantic meaning. Therefore, many feature-reconstruction-based methods have been proposed. As an example, DFR \cite{DFR} directly reconstructs the multiple hierarchical regional features extracted by a pre-trained convolutional neural network (CNN) and leverages the feature reconstruction errors to perform anomaly segmentation.\\
\indent Nonetheless, existing feature-reconstruction-based methods suffer from trivial solutions or shortcut learning, where the model simply reconstructs a copy of input features without considering global semantic information, This can lead to mis-inspections, as shown in Fig. \ref{fig:global-defect}. To address this challenge, we propose a template-based feature aggregation network (TFA-Net) that enables the model to reconstruct features by considering global semantic information, rather than simply copying the input data. Firstly, considering the similarity of features among normal samples, we select a normal sample as a fixed template image. Next, TFA-Net utilizes a pre-trained CNN to extract multiple hierarchical levels of semantic features from both the input image and the template image. The input features are used as the reconstruction target, rather than the low-level image pixels with low semantic information. Additionally, a novel template-based feature aggregation mechanism (TFAM) is proposed to address the issue of shortcut learning in existing reconstruction methods. The objective of TFAM is to aggregate the input features onto the normal template features, thereby transforming a trivial feature reconstruction task into a nontrivial and meaningful feature aggregation task. In the selection of feature aggregation networks, we opted for the vision transformer (ViT) \cite{Vit} network over the CNN network. As shown in Fig. \ref{fig:VIT-CNN-COM}, the CNN network incorporates two inductive biases: locality and translation equivariance, which limit its ability to perform feature aggregation on parts with different orientations. In contrast, ViT lacks such inductive biases and is better suited for modeling global information, thereby facilitating more effective feature aggregation. Subsequently, the template features, which have already incorporated input features, are fed into the feature detail refinement module (FDRM) to obtain the reconstructed features. Finally, the feature residual between input features and reconstructed features can be utilized as a reliable anomaly score.\\
\indent In general, the major contributions can be listed as follows:
\begin{itemize}
    \item TFA-Net leverages a pre-trained CNN to extract image features with diverse receptive fields. These features undergo multi-scale fusion, with the fused features serving as the reconstruction target. Due to their enriched semantic information, the fused features enable the detection of defects of varying sizes.
    \item We present a novel template-based feature aggregation mechanism (TFAM). TFAM aggregates normal features from the input onto the normal template feature, efficiently filtering out abnormal features, and resulting in meaningful reconstruction rather than simple data replication. This effectively addresses the problem of shortcut learning in existing reconstruction methods.
    \item The dual-mode anomaly segmentation method leverages both input features and reconstructed features for accurate defect localization. Notably, it employs two distinct similarity metrics, namely Euclidean distance and cosine similarity, to measure the similarity between features. This approach significantly improves the robustness of anomaly detection.
    \item The proposed TFA-Net demonstrates enhanced anomaly detection performance on the MVTec AD \cite{MVTEC} benchmark, achieving the area under the receiver operating characteristic curve (AUROC) of 98.7\% for anomaly detection and 98.3\% for anomaly segmentation across all 15 categories. Furthermore, our proposed approach also exhibits outstanding detection performance on the recently released MVTec LOCO AD \cite{mvtecloco} dataset.
\end{itemize}

The structure of the remaining article is as follows: Section \MakeUppercase{\romannumeral2} provides an overview of related work on VAD, Section \MakeUppercase{\romannumeral3} presents a detailed description of our proposed method, Section \MakeUppercase{\romannumeral4} reports the experimental results, and the final section summarizes the entire article.

\section{Related Work}
Visual anomaly detection is a fundamental and important problem in the field of computer vision, with wide applications in industrial inspection. It has attracted a large number of scholars to conduct research in this area. In general, anomaly detection methods can be mainly classified into two categories: embedding-based methods and reconstruction-based methods.
\subsection{Embedding-based Methods}
Embedding-based methods establish a compressed discriminative space based on a large amount of normal data, where the difference between anomalous data and the normal center is significant. Hence, the embedding distance between them can be used as a reliable anomaly score. Deep SVDD \cite{DeepSVDD} is capable of establishing normal clustering at the image level, thereby enabling anomaly detection. On the other hand, Patch SVDD \cite{PatchSVDD} can establish multiple normal clusters at the patch level, enabling precise localization of anomalies. Bergmann \textit{et al}. \cite{ST} employed the prior knowledge of a pre-trained network to train a student framework, with the regression error between them utilized for defect detection and localization. Padim \cite{PaDiM} and GCPF \cite{GCPF} model normal features using multivariate Gaussian distributions, and calculate anomaly scores by computing the Mahalanobis distance between the test features and the Gaussian distribution. SPADE \cite{SPADE} and PatchCore \cite{PacthCore} employ a memory bank to store normal features and utilize the distance between the test features and the most similar normal feature in the memory bank as the anomaly criterion. MBPFM \cite{MBPFM} leverages a multi-level bidirectional feature mapping for accurate anomaly localization. While the aforementioned methods have demonstrated promising results in the anomaly detection field, their practical utility in industrial settings is constrained by the significant online memory requirements and low inference speed they entail.
\begin{figure*}
    \centering
    \includegraphics[width=\textwidth]{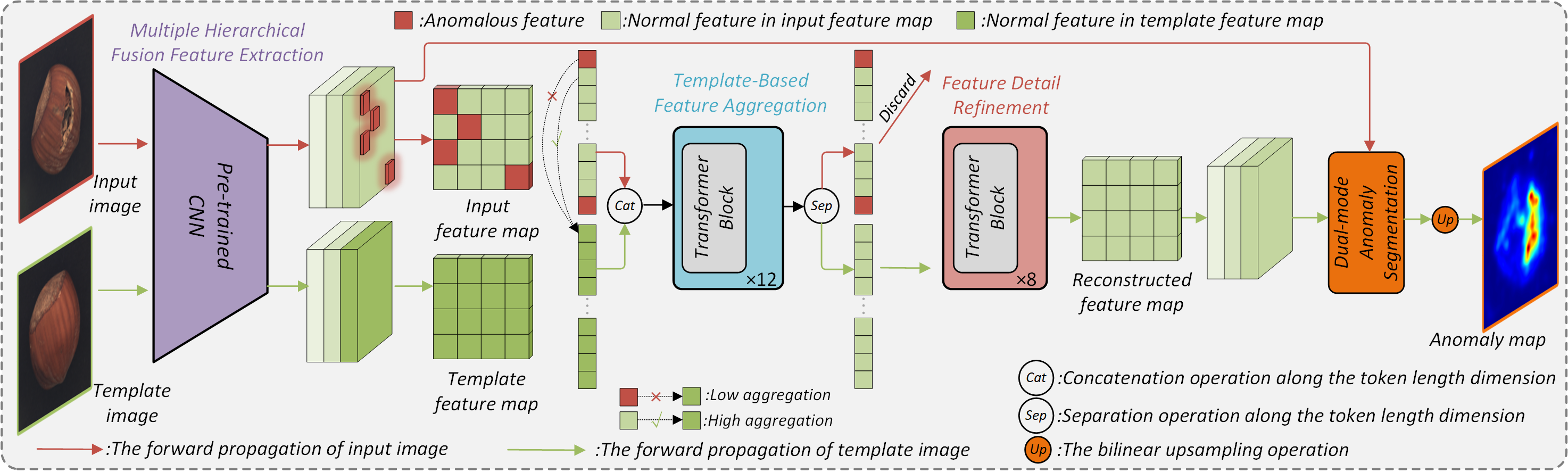}
    \caption{The overall architecture of TFA-Net approach. The workflow of TFA-Net can be divided into four stages: multiple hierarchical fusion feature extraction, template-based feature aggregation mechanism (TFAM), feature detail refinement module (FDRM), and dual-mode anomaly segmentation. Firstly, TFA-Net employs a pre-trained CNN to extract multi-scale features. Subsequently, TFAM is utilized to filter out anomalous features while retaining normal features. Further reconstruction of the features is performed using FDRM, resulting in reconstructed features. Finally, the defect regions are localized by leveraging both the reconstructed features and the input features.}
    \label{fig:network}
\end{figure*}
\subsection{Reconstruction-based Methods}
Reconstruction-based methods model the distribution of normal data by encoding and decoding it during training, and use the reconstruction error as an anomaly criterion during testing. Autoencoders (AE) \cite{AE} are a well-known type of reconstruction model, but their high generalization ability can be a drawback, as they tend to accurately reconstruct even defective data, leading to the potential for false negatives in anomaly detection. To address this issue, numerous improved AE-based models have been proposed. For instance, Bergmann \textit{et al}. \cite{AE-SSIM} introduced AE-SSIM, a model that enables a more targeted focus on image structural information, thus enhancing the model's capability to detect anomalies. Dong \textit{et al}. \cite{MemAE} proposed a sparse memory mechanism to alleviate the generalization limitation of vanilla AE. MSCDAE \cite{MSCDAE} employs denoising proxy tasks at multiple scales to enhance the performance of the model, while MSFCAE \cite{MSFCAE} performs feature clustering at multiple levels to constrain the distribution of latent features. RIAD \cite{RIAD} is an approach that leverages both U-net \cite{U-net} and random masking strategy to transform the reconstruction task into an image inpainting task. NDP-Net \cite{NDP-Net} enhances the discriminative power of the model by utilizing artificial defect samples, thereby improving its performance in detecting unknown defects. Recently, Generative Adversarial Networks (GANs) \cite{GANs} have attracted widespread attention due to their powerful generative capabilities. Schlegl \textit{et al}. \cite{AnoGan} proposed AnoGAN for detecting defects by learning the distribution of normal data with a GAN. However, AnoGAN lacks the mapping from the image domain to the feature domain, thus many methods \cite{f-anogan, OCGAN, Ganomaly, skip-ganomaly} have been proposed to address this issue. All of the aforementioned methods focus on image-level reconstruction. DFR \cite{DFR} employs the features extracted by a pre-trained network as the reconstruction target. ST-MAE \cite{STMAE} achieves good detection accuracy by performing feature reconstruction from the perspective of complementary feature transformation. UTRAD \cite{chen2022utrad} introduces a U-shaped Transformer for feature reconstruction, effectively reducing computational complexity and achieving higher detection and segmentation accuracy. In general, the aforementioned reconstruction methods have all experienced the problem of shortcut learning, which results in defects being perfectly reconstructed.

\begin{figure}[!t]
    \centering
    \includegraphics[width=83mm]{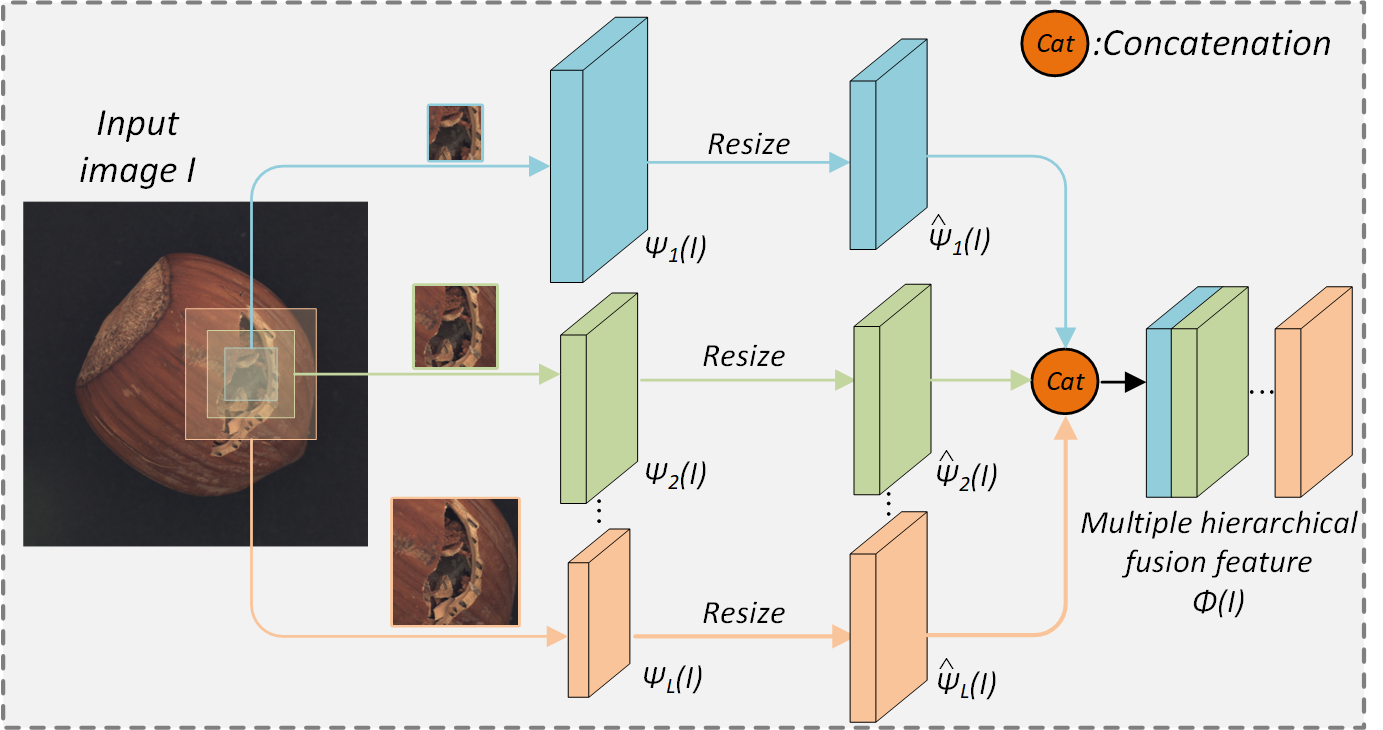}
    \caption{Illustration of multiple hierarchical fusion feature extraction, {which encompasses the process of resizing feature maps from various scales to a uniform size and subsequently concatenating them in the channel dimension to obtain multi-level fused features.}}
    \label{fig:multi-feature}
\end{figure}
\section{The TFA-Net Methodology}
In this section, we present our proposed method in detail. Firstly, we provide an overview of the overall framework of our network. Next, we describe the extraction of multi-level semantic features. Then, we introduce the template-based feature aggregation mechanism and the feature detail optimization module. Finally, we explain the training and testing process of our model.
\subsection{Overall Architecture}
Fig. \ref{fig:network} illustrates the overall architecture of TFA-Net, a hybrid model combining CNNs and ViT \cite{Vit}. Firstly, {a pre-trained CNN is employed} to extract multi-level input features and template features, which are then respectively scaled to the same size and concatenated. {Due to the high similarity between the input normal features and normal template features, the normal input features readily aggregate onto the template features, manifesting high aggregation. In contrast, defect features and normal template features exhibit low similarity, rendering defect features less likely to aggregate onto the template features, thereby displaying low aggregation.} Therefore, the TFAM aggregates the input normal features onto the template features, effectively filtering out anomalous features and generating a meaningful and challenging task. Then, we discard the input features and perform feature refinement on the template features that have already aggregated the input features, resulting in reconstructed features. Finally, we utilize a dual-mode anomaly segmentation method to accurately detect and locate anomalies.

\begin{figure}
    \centering
    \includegraphics[width=83mm]{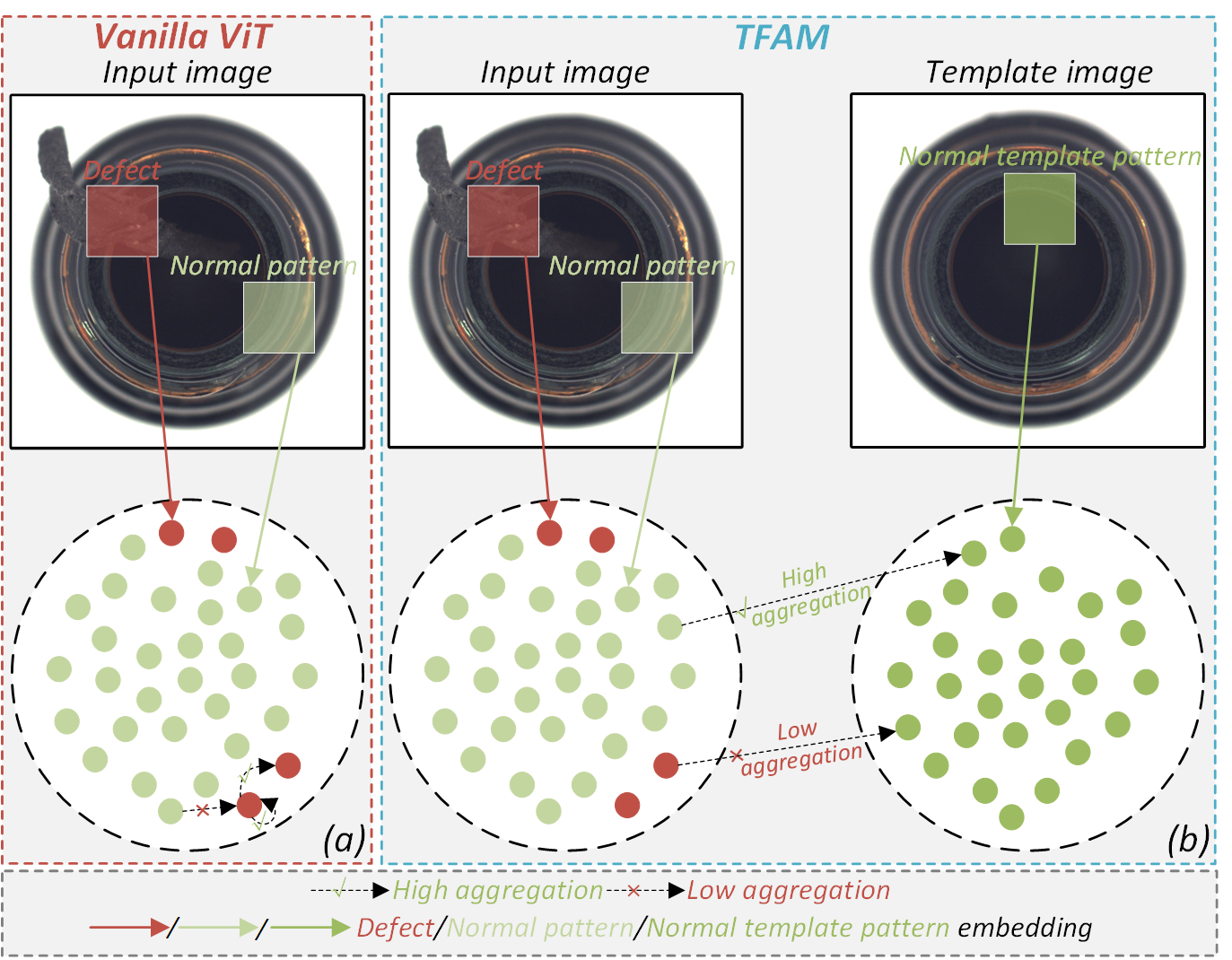}
    \caption{Comparison between Vanilla ViT and Template-based feature aggregation mechanism (TFAM). (a) Vanilla ViT mechanism. {In the ViT mechanism, defect features exhibit the highest similarity with themselves, causing defect features to self-aggregate and consequently resulting in a perfect reconstruction of defects.} (b) TFAM. {In the TFAM, defect features are dissimilar to template features, making it challenging for them to aggregate onto template features.}}
    \label{fig:TFAM}
\end{figure}
\subsection{Multiple Hierarchical Fusion Feature Extraction}
Current existing reconstruction-based methods perform image-level reconstruction and utilize pixel-wise differences between input and reconstructed images as anomaly scores. However, this leads to two problems: 1) difficulty in accurately reconstructing image details. 2) pixel-wise differences between input and reconstructed images are semantically meaningless and cannot serve as a reliable anomaly criterion. Therefore, in this paper, we utilize the features extracted by a pre-trained CNN $\psi$ on the ImageNet \cite{ImageNet} dataset as the reconstruction target. As shown in Fig. \ref{fig:multi-feature}, given an image $I$, we can obtain different hierarchical feature maps $\{\psi_1(I),\psi_2(I),\cdots, \psi_L(I)\}$ through the pre-trained CNN $\psi$, where $\psi_l(I)\in R^{H^l\times W^l \times C^l}$. Different hierarchical feature maps have different receptive fields, with increasing receptive fields leading to less detailed information but richer semantic information in the feature maps. Therefore, in order to strike a balance between detail and semantic information, we scale the features from different levels to the same size and concatenate them along the channel dimension to obtain a multiple hierarchical fusion feature $\phi(I)$.
\begin{equation}
    \hat{\psi}_l(I)=Resize(\psi_l(I)) 
\end{equation}
\vspace{-2.0em}
\begin{equation}
       \phi(I)=U(\hat{\psi}_1(I),\hat{\psi}_2(I),\cdots,\hat{\psi}_L(I))
\end{equation}
where $\phi(I)\in R^{H\times W \times C}$, $Resize$ denotes the scaling operation, and $U$ represents the concatenation operation along the channel dimension. {In the TFA-Net method, we employ Wide-Resnet50 \cite{wideresnet} as the feature extractor, where feature maps from the 1st to 4th layers are used for fusion, resulting in the final multi-scale feature map with size of $64\times 64$ and a feature dimension of 1856.}\\
\indent Compared to image-level reconstruction, using this multi-level fusion feature as the reconstruction target can effectively improve the inspection performance of the model.

\subsection{Template-based Feature Aggregation Mechanism and Feature Detail Refinement Module}
\begin{figure}[!t]
    \centering
    \includegraphics{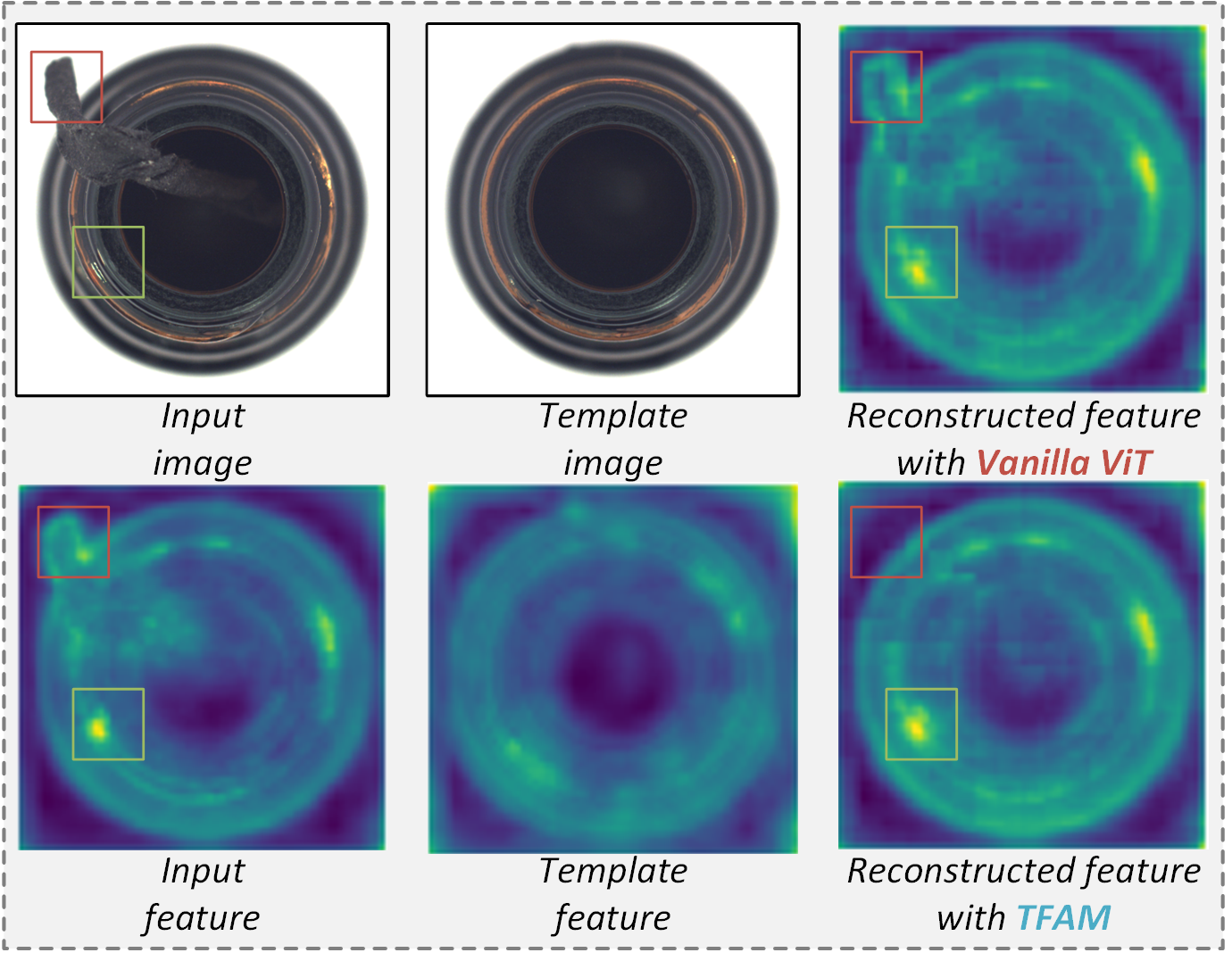}
    \caption{The effect of TFAM. {The features reconstructed with Vanilla ViT still exhibit defects. Conversely, those reconstructed with TFAM not only rectify the defects but also retain information about normal patterns.} The red box indicates the defective region, while the green box implies the normal pattern.}
    \label{fig:TFAM-1}
\end{figure}

\begin{figure}[!t]
    \centering
    \includegraphics{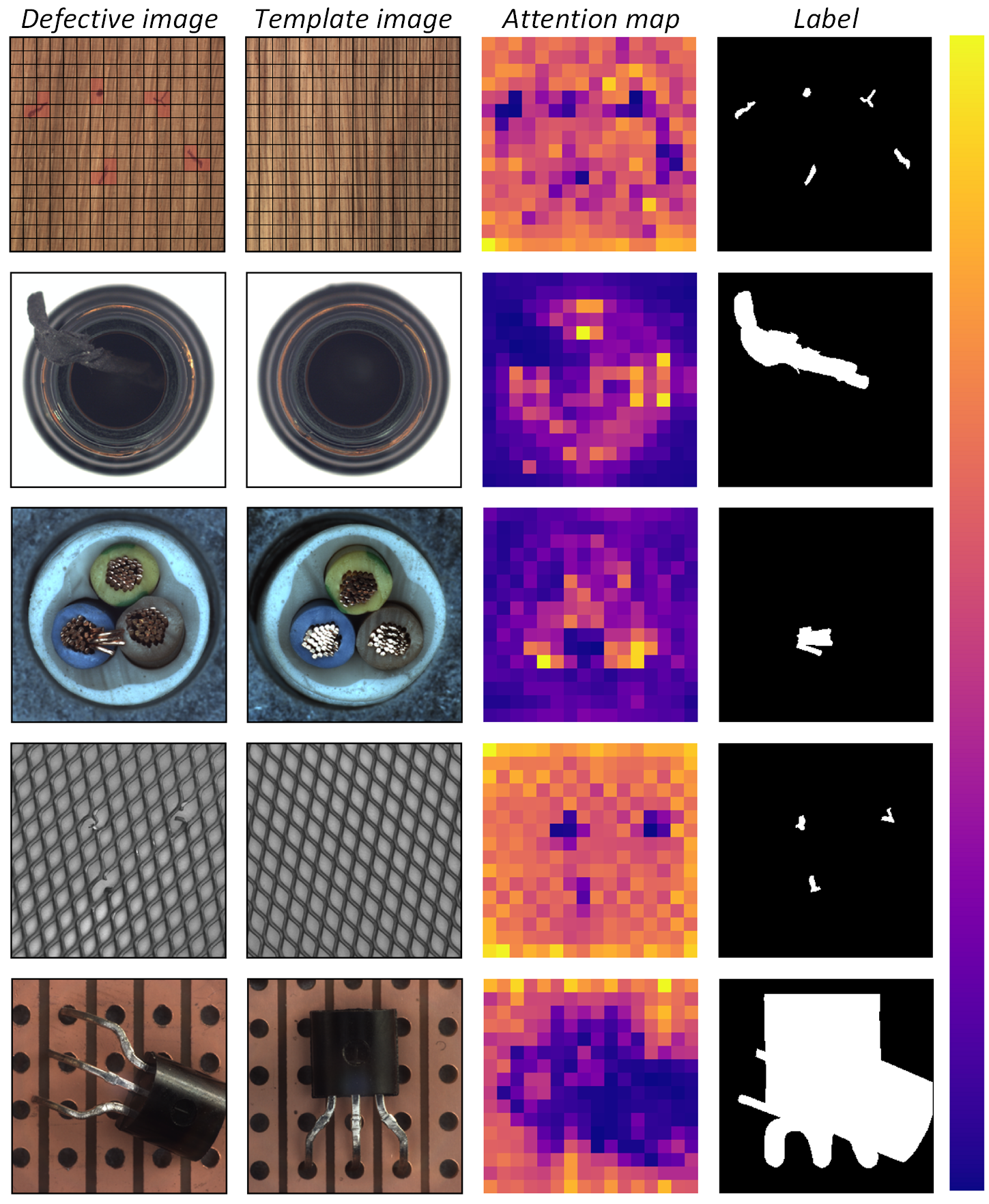}
    \caption{{The visualization of attention maps in TFAM. The attention map is obtained by calculating the sum of similarities between each input feature patch and all template feature patches.
}}
    \label{fig:TFAM-attention}
\end{figure}

Both feature reconstruction and image reconstruction methods suffer from the problem of shortcut learning, which results in defects being perfectly reconstructed and thus leading to false positive detection. To address this issue, we propose the novel template-based feature aggregation mechanism (TFAM).\\
\indent As shown in Fig. \ref{fig:network}, we innovatively introduce a normal template image and utilize a pre-trained CNN $\psi$ to obtain the multi-level fused features of the input image $I$ and the template image $I_T$, which are respectively referred to as input feature map $\phi(I)$ and template feature map $\phi(I_T)$. With the same projection heads, we utilized the patch size $K$ to embed $\phi(I)$ and $\phi(I_T)$ into sequences of token embeddings $E_{\phi(I)}$ and $E_{\phi(I_T)}$ of length $N$ and dimension $D$.
\begin{equation}
     E_{\phi(I)}=[E_{\phi(I)}^1, E_{\phi(I)}^2, \cdots, E_{\phi(I)}^N]+E_{pos}
\end{equation}
\vspace{-2.0em}
\begin{equation}
    E_{\phi(I_T)}=[E_{\phi(I_T)}^1, E_{\phi(I_T)}^2, \cdots, E_{\phi(I_T)}^N]+E_{pos}
\end{equation}
where $E_{\phi(I)}^i,E_{\phi(I_T)}^i\in R^{D}$, $N=\frac{H}{K}\times \frac{W}{K}$, and $E_{pos}$ represents the positional embeddings. {The incorporation of positional information plays a crucial role in enhancing the performance of TFA-Net. Previous studies, as referenced in \cite{wanqiancase,gudovskiy2022cflow}, have noted that the incorporation of positional encoding enhances the model's capacity to capture spatial relationships, thereby yielding more accurate results in anomaly detection.}\\
\indent As depicted in Fig. \ref{fig:TFAM}(a), the vanilla ViT \cite{Vit} directly input $E_{\phi(I)}$ into the transformer blocks (TBs), which indeed lead to the problem of abnormal features being more likely to be correlated with themselves or adjacent abnormal features. As shown in Fig. \ref{fig:network} and Fig. \ref{fig:TFAM}(b), our proposed TFAM concatenates the token embeddings $E_{\phi(I)}$ and $E_{\phi(I_T)}$ to obtain $E_{cat}\in R^{2N\times D}$, which is then input into the TBs. In this way, during the self-attention process, the normal features in $E_{\phi(I)}$ can be aggregated into $E_{\phi(I_T)}$ since the normal features in $E_{\phi(I)}$ and $E_{\phi(I_T)}$ are similar, while the abnormal features in $E_{\phi(I)}$ are difficult to be aggregated into $E_{\phi(I_T)}$ due to the absence of abnormal features in $E_{\phi(I_T)}$. Consequently, the TFAM effectively filters out abnormal features. After the integration of $E_{\phi(I)}$ and $E_{\phi(I_T)}$ using TFAM, the normal semantic information from $E_{\phi(I)}$ is already aggregated in $E_{\phi(I_T)}$. Therefore, we discard $E_{\phi(I)}$ and retain $E_{\phi(I_T)}$,  which is further refined to obtain reconstructed token embedding $\hat{E}_{\phi(I)}$ using feature detail refinement module (FDRM), which consists of a serial of TBs. The final reconstructed feature map $\hat{\phi}(I)$ can be obtained through a linear projection head and the reconstructed token embeddings  $\hat{E}_{\phi(I)}$.\\
\indent Fig. \ref{fig:TFAM-1} illustrates the superiority of TFAM over the vanilla ViT. The vanilla ViT can perfectly reconstruct the defective features, whereas TFAM employs normal template features to filter out the defective ones. Due to the high similarity between the normal features in the input and the normal template features, the normal features in the input are aggregated into the normal template features. Conversely, the defective features exhibit low similarity with the normal template features, leading to their filtration and the subsequent generation of a reconstructed feature map free from anomalies.\\
\indent {The effectiveness of TFAM is further validated from the perspective of attention map visualization in Fig. \ref{fig:TFAM-attention}. The attention map is obtained by calculating the sum of similarities between each input feature patch and all template feature patches. We observe that the attention map assigns smaller values to defective regions and larger values to normal regions. This observation indicates that defect features are less prone to aggregate into the template features, whereas normal features tend to aggregate more readily into the template features. This finding aligns entirely with our previous explanations.}\\
\indent Although the selection of 
template images appears to be crucial for TFAM, the ablative experiments conducted in Section \ref{tempate-ablate} demonstrate the robustness of TFA-Net's detection performance to template image selection. Different template images do not significantly affect the detection performance of TFA-Net. This is because, even though these template images may appear completely different in appearance, when divided into patches, patches from different positions can still correspond to each other. Furthermore, the ViT model lacks translational equivariance and local inductive biases, enabling it to establish mutual correspondences among patches from different positions. Therefore, in this study, we selected the first normal image from the training dataset as the fixed template image.\\
\indent In summary, our proposed TFAM, together with FDRM, effectively addresses the issue of trivial solution in existing feature reconstruction methods and leads to a more meaningful feature reconstruction task. The feature reconstruction by TFAM and FDRM is summarized in Algorithm \ref{alg:TFAM}.

\begin{algorithm}[!t]
    \caption{Feature reconstruction by TFAM and FDRM}
    \label{alg:TFAM}
     \KwIn{Input feature $\phi(I)$, Template feature $\phi(I_T)$}
    \KwOut{Reconstructed feature $\hat{\phi}(I)$}
    Obtain patch embeddings:
    $E_{\phi(I)}, E_{\phi(I_T)} = \textit{\textbf{Proj.}}(\phi(I)),\textit{\textbf{Proj.}}(\phi(I_T))$ \\
    Concatenate $E_{\phi(I)}$ and $E_{\phi(I_T)}$: $E_{cat} = \{E_{\phi(I)}, E_{\phi(I_T)}\}$\\
    Template-based feature aggregation:
          $E_{cat}=\textit{\textbf{Trans.-Blocks}}(E_{cat})$\\
    Split $E_{cat}$, discard $E_{\phi(I)}$, and retain $E_{\phi(I_T)}$ \\
    Feature detail refinement:
    $\hat{E}_{\phi(I)}=\textit{\textbf{Trans.-Blocks}}(E_{\phi(I_T)})$\\
    Obtain reconstructed feature:
    $\hat{\phi}(I)=\textit{\textbf{Inverse-Proj.}}(\hat{E}_{\phi(I)})$\\
    return  $\hat{\phi}(I)$
\end{algorithm}
\begin{figure*}[h]
    \centering
    \includegraphics[width=165mm]{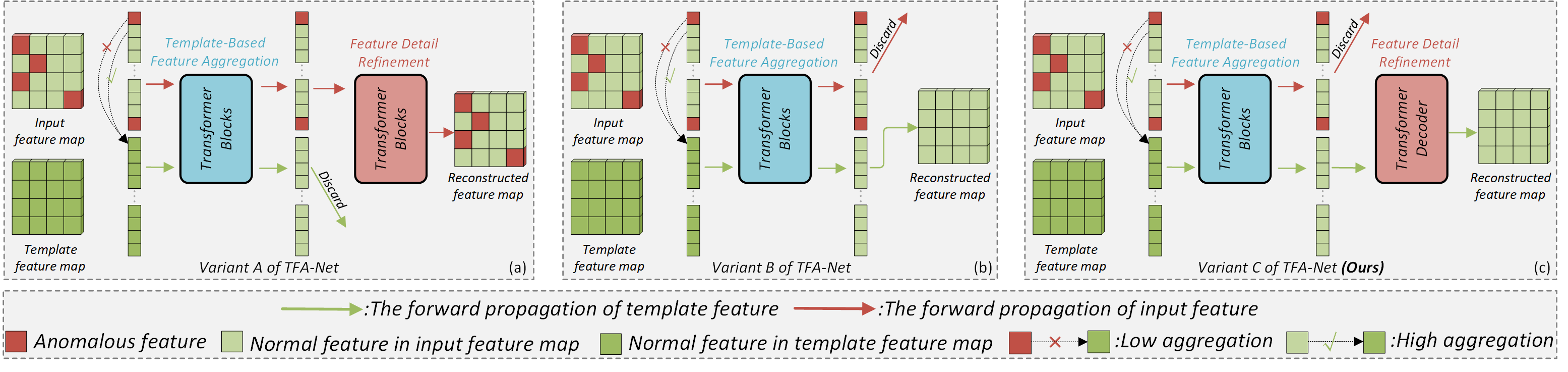}
    \caption{{Three variants of TFA-Net. (a) Variant A: TFA-Net model variant that discards template features and retains input features. (b) Variant B: TFA-Net model variant without FDRM. (c) Variant C (\textbf{Ours}): TFA-Net model variant that discards input features and retains template features.}}
    \label{fig:TFA-Net-variation}
\end{figure*}
\begin{figure}[h]
    \centering
    \includegraphics{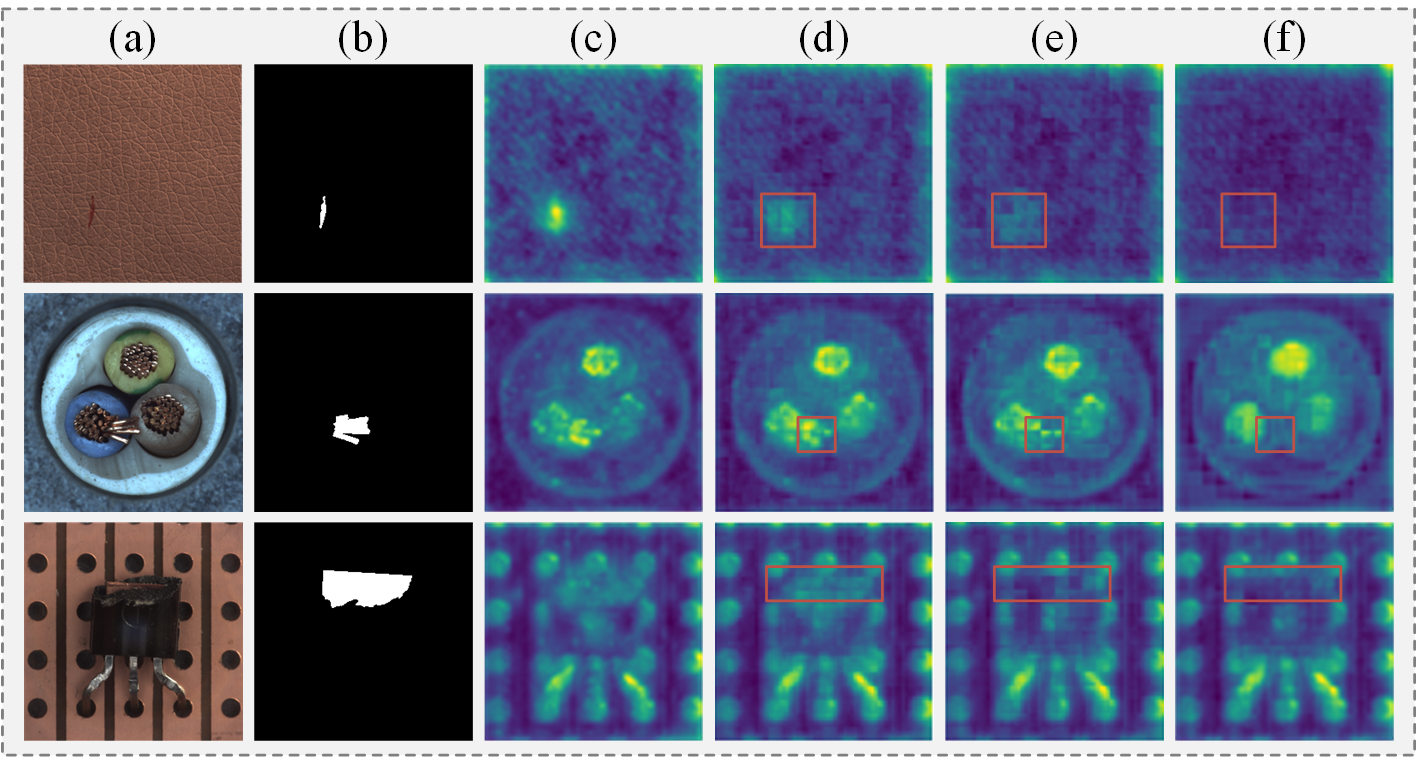}
    \caption{{The reconstructed results of the three TFA-Net model variants in Fig. \ref{fig:TFA-Net-variation}. (a) The defective image. (b) The corresponding label. (c) The defective feature map. (d) The reconstructed result of variant A in Fig. \ref{fig:TFA-Net-variation}. (e) The reconstructed result of variant B in Fig. \ref{fig:TFA-Net-variation}. (f) The reconstructed result of variant C in Fig. \ref{fig:TFA-Net-variation}.}}
    \label{fig:example_var}
\end{figure}
\subsection{{Discussion on TFAM and FDRM}}
{In this subsection, we delve into a more comprehensive discussion of TFAM and FDRM, primarily to address two key issues. Firstly, we explain why we choose to discard the input features after TFAM rather than the template features. Secondly, given TFAM's capability to effectively filter defect characteristics, we explore the specific role of FDRM. These considerations led us to devise two variants of TFA-Net, as illustrated in Fig. \ref{fig:TFA-Net-variation}.}
\subsubsection{{The reason for discarding the input features after passing through TFAM}}{If we discard template features and retain the input features after TFAM, the defect features cannot be restored to normal features due to the high similarity between the defect features and themselves. Therefore, the defect features are perfectly reconstructed. On the other hand, if we discard the input features and retain the template features after TFAM, due to the low similarity between defect features and template features, defect features cannot be aggregated into template features. Consequently, defect features cannot be reconstructed. Columns (d) and (f) in Figure \ref{fig:example_var} validate our previously stated explanation.}
\subsubsection{{The specific effect of FDRM}}{
While TFAM possesses the capability to filter defect features, it is noteworthy that defect features and template features consistently reside within the same latent feature space in TFAM. Consequently, even if defect features and template features exhibit low similarity, there exists a potential for a small subset of defect features to be aggregated into the template features. Therefore, we individually input the template features processed by TFAM into FDRM, aiming to refine and repair the limited defect features within the template features. Columns (e) and (f) in Fig.\ref{fig:example_var} validate the efficacy of FDRM in this regard.}
\subsection{Training and Testing Procedures}
\subsubsection{Training procedure}
Compared to DFR \cite{DFR}, our approach utilizes two similarity metrics, namely Euclidean distance and cosine similarity, to compare features. Consequently, we use a joint optimization approach that involves both the Euclidean distance loss $L_{euc}$ and cosine similarity loss $L_{cos}$ in the entire optimization process of the TFA-Net. Given an input feature $\phi(I)\in R^{H\times W \times C}$, TFA-Net reconstructs it into feature $\hat{\phi}(I)\in R^{H\times W \times C}$, and the joint loss function is defined as follows:
\begin{equation}
     L_{euc}=\underset{I \sim P_{I}}{\mathbb{E}}\left [\big\|\phi(I)-\hat{\phi}(I)\big \|_2\right ]
\end{equation}
\vspace{-1.0em}
\begin{equation}
     L_{cos}=\underset{I \sim P_{I}}{\mathbb{E}}\left [1-\frac{\phi(I)\cdot\hat{\phi}(I)}{\big\|\phi(I)\big\| \times\big\|\hat{\phi}(I)\big\|}\right ]
\end{equation}
\vspace{-1.0em}
\begin{equation}
        L_{rec}=w_{euc}L_{euc}+w_{cos}L_{cos}
\end{equation}
where ${\big\|\cdot\big\|}_2$, $\cdot$, and $\big\|\cdot\big\|$ represent the $L_2$ norm, inner product, and modulus length, respectively. {$w_{euc}$} and $w_{cos}$ are the weights assigned to the two types of losses. In this study, we set $w_{euc}$ and $w_{cos}$ to 1 and 5, respectively.
\subsubsection{Testing procedure}
\begin{figure}[!t]
    \centering
    \includegraphics{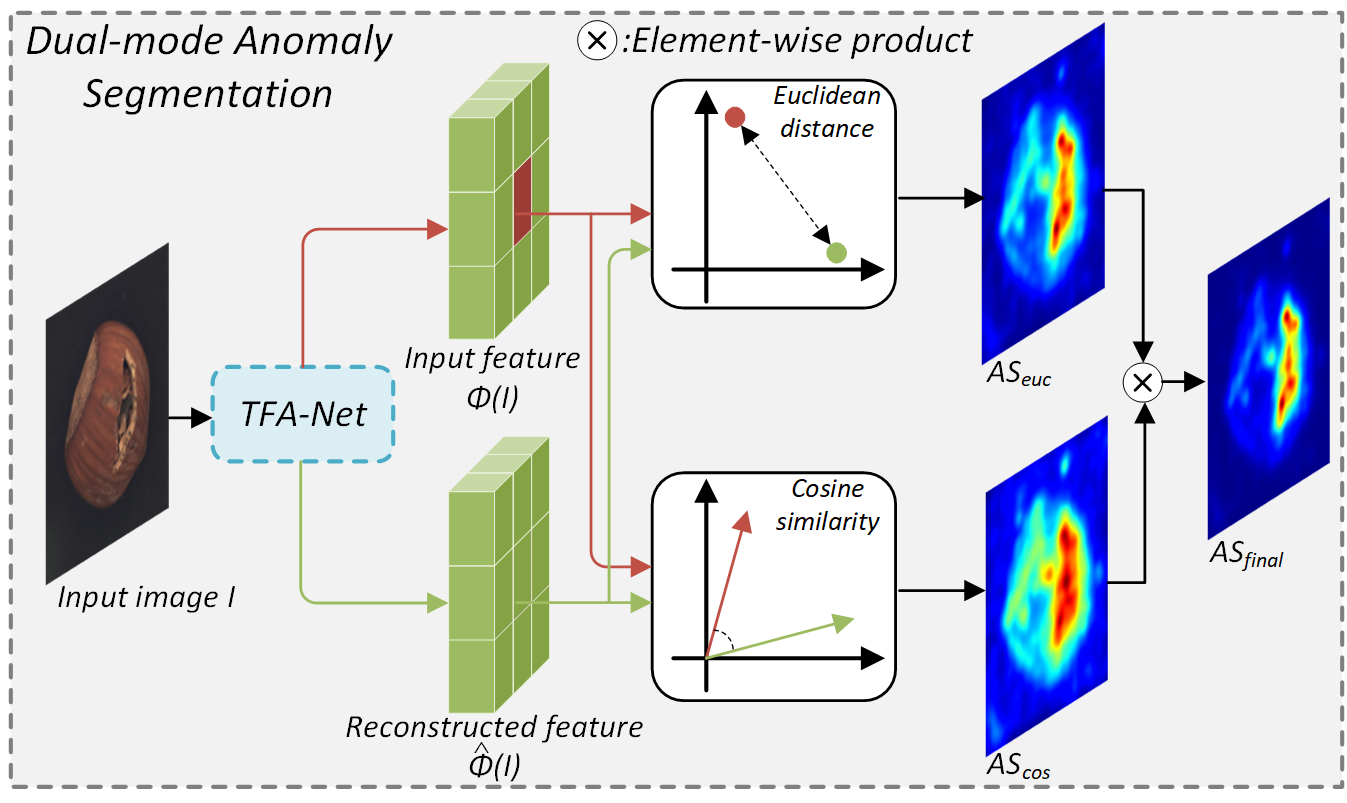}
    \caption{Schematic diagram of dual-mode anomaly segmentation, {which employs both Euclidean distance and cosine similarity to jointly measure the differences between input and reconstructed features.}}
    \label{fig:dual-mode}
\end{figure}

After training, TFA-Net can be used for practical industrial anomaly detection. As shown in Fig. \ref{fig:dual-mode}, given an anomalous image, TFA-Net is able to filter out the anomalous features and retain the normal features, thereby obtaining a reconstructed feature map that eliminates anomalies. Then, we use a dual-mode anomaly segmentation method for defect detection and localization:
\begin{equation}
 AS_{final}=\underbrace{\big\|\phi(I)-\hat{\phi}(I)\big\|_2}_{AS_{euc}} \otimes \underbrace{(1-\frac{\phi(I)\cdot\hat{\phi}(I)}{\big\|\phi(I)\big\|\times\big\|\hat{\phi}(I)\big\|})}_{AS_{cos}}
\end{equation}
where $\otimes$ represents the element-wise product. Finally, the $AS_{final}$ is upsampled to match the size of the input image through interpolation. A Gaussian filter is then applied to $AS_{final}$ for smoothing. Similar to \cite{multilevel-recon}, we calculate the standard deviation of the smoothed $AS_{final}$ as image-level anomaly score.
\begin{table*}[!t,width=1.0\textwidth,cols=4,pos=h]
\centering
\caption{The image/pixel level AUROC results on MVTec AD dataset}
\label{Table:mvtec}
\begin{threeparttable}
\resizebox{\textwidth}{!}{
\begin{tabular}{cccccccccccc}
\Xhline{1.5pt}
{\color[HTML]{000000} }                           & \multicolumn{5}{c}{{\color[HTML]{000000} Reconstruction-based methods}}                                                                                                      & {\color[HTML]{000000} } & \multicolumn{4}{c}{{\color[HTML]{000000} Embedding-based methods}}                                                                         & {\color[HTML]{000000} }                                   \\ \cline{2-6} \cline{8-11}
\multirow{-2}{*}{{\color[HTML]{000000} Category}} & {\color[HTML]{000000} TrustMAE}  & {\color[HTML]{000000} RIAD}      & {\color[HTML]{000000} DFR}       & {\color[HTML]{000000} Draem}     & {\color[HTML]{000000} Intra}     & {\color[HTML]{000000} } & {\color[HTML]{000000} SPADE}     & {\color[HTML]{000000} Pacth SVDD} & {\color[HTML]{000000} GCPF}      & {\color[HTML]{000000} MBPFM}     & \multirow{-2}{*}{{\color[HTML]{000000} \textbf{TFA-Net}}} \\ \hline
{\color[HTML]{000000} Carpet}                     & {\color[HTML]{000000} 97.4/98.5} & {\color[HTML]{000000} 84.2/94.2} & {\color[HTML]{000000} -/97.0}    & {\color[HTML]{000000} 97.0/95.5} & {\color[HTML]{000000} 98.9/\underline{99.2}} & {\color[HTML]{000000} } & {\color[HTML]{000000} 92.8/97.5} & {\color[HTML]{000000} 92.9/92.6}  & {\color[HTML]{000000} -/98.9}    & {\color[HTML]{000000} \textbf{100}/\underline{99.2}}  & {\color[HTML]{000000} \underline{99.8}/\textbf{99.3}}                          \\
{\color[HTML]{000000} Grid}                       & {\color[HTML]{000000} 99.1/97.5} & {\color[HTML]{000000} 99.6/96.3} & {\color[HTML]{000000} -/98.0}    & {\color[HTML]{000000} \underline{99.9}/ \textbf{99.3}} & {\color[HTML]{000000} \textbf{100}/98.8}  & {\color[HTML]{000000} } & {\color[HTML]{000000} 47.3/93.7} & {\color[HTML]{000000} 94.6/96.2}  & {\color[HTML]{000000} -/97.8}    & {\color[HTML]{000000} 98.0/98.8} & {\color[HTML]{000000} 99.7/\underline{99.0}}                          \\
{\color[HTML]{000000} Leather}                    & {\color[HTML]{000000} 95.1/98.1} & {\color[HTML]{000000} \textbf{100}/\underline{99.4}}  & {\color[HTML]{000000} -/98.0}    & {\color[HTML]{000000} \textbf{100}/98.6}  & {\color[HTML]{000000} \textbf{100}/\textbf{99.5}}  & {\color[HTML]{000000} } & {\color[HTML]{000000} \underline{95.4}/97.6} & {\color[HTML]{000000} 90.9/97.4}  & {\color[HTML]{000000} -/99.3}    & {\color[HTML]{000000} \textbf{100}/\underline{99.4}}  & {\color[HTML]{000000} \textbf{100}/\underline{99.4}}                           \\
{\color[HTML]{000000} Tile}                       & {\color[HTML]{000000} 97.3/82.5} & {\color[HTML]{000000} 98.7/89.1} & {\color[HTML]{000000} -/87.0}    & {\color[HTML]{000000} \underline{99.6}/\textbf{99.2}} & {\color[HTML]{000000} 98.2/94.4} & {\color[HTML]{000000} } & {\color[HTML]{000000} 96.5/87.4} & {\color[HTML]{000000} 97.8/91.4}  & {\color[HTML]{000000} -/96.1}    & {\color[HTML]{000000} \underline{99.6}/\underline{96.2}} & {\color[HTML]{000000} \textbf{100}/95.9}                           \\
{\color[HTML]{000000} Wood}                       & {\color[HTML]{000000} \textbf{99.8}/92.6} & {\color[HTML]{000000} 93.0/85.8} & {\color[HTML]{000000} -/94.0}    & {\color[HTML]{000000} 99.1/\textbf{96.8}} & {\color[HTML]{000000} 97.5/88.7} & {\color[HTML]{000000} } & {\color[HTML]{000000} 95.8/88.5} & {\color[HTML]{000000} 96.5/90.8}  & {\color[HTML]{000000} -/95.1}    & {\color[HTML]{000000} 99.5/95.6} & {\color[HTML]{000000} \underline{99.7}/\underline{95.8}}                          \\
{\color[HTML]{000000} Bottle}                     & {\color[HTML]{000000} 97.0/93.4} & {\color[HTML]{000000} \underline{99.9}/98.4} & {\color[HTML]{000000} -/97.0}    & {\color[HTML]{000000} 99.2/\textbf{99.1}} & {\color[HTML]{000000} \textbf{100}/97.1}  & {\color[HTML]{000000} } & {\color[HTML]{000000} 97.2/98.4} & {\color[HTML]{000000} 98.6/98.1}  & {\color[HTML]{000000} -/97.5}    & {\color[HTML]{000000} \textbf{100}/98.4}  & {\color[HTML]{000000} \textbf{100}/\underline{98.5}}                           \\
{\color[HTML]{000000} Cable}                      & {\color[HTML]{000000} 85.1/92.9} & {\color[HTML]{000000} 81.9/84.2} & {\color[HTML]{000000} -/92.0}    & {\color[HTML]{000000} 91.8/94.7} & {\color[HTML]{000000} 70.3/91.0} & {\color[HTML]{000000} } & {\color[HTML]{000000} 84.8/\underline{97.2}} & {\color[HTML]{000000} 90.3/96.8}  & {\color[HTML]{000000} -/95.7}    & {\color[HTML]{000000} \textbf{98.8}/96.7} & {\color[HTML]{000000} \underline{96.5}/\textbf{98.1}}                          \\
{\color[HTML]{000000} Capsule}                    & {\color[HTML]{000000} 78.8/87.4} & {\color[HTML]{000000} 88.4/92.8} & {\color[HTML]{000000} -/\textbf{99.0}}    & {\color[HTML]{000000} \textbf{98.5}/94.3} & {\color[HTML]{000000} 86.5/97.7} & {\color[HTML]{000000} } & {\color[HTML]{000000} 91.0/\textbf{99.0}} & {\color[HTML]{000000} 76.7/95.8}  & {\color[HTML]{000000} -/97.2}    & {\color[HTML]{000000} 94.5/\underline{98.3}} & {\color[HTML]{000000} \underline{94.8}/\textbf{99.0}}                          \\
{\color[HTML]{000000} Hazelnut}                   & {\color[HTML]{000000} \underline{98.5}/98.5} & {\color[HTML]{000000} 83.3/96.1} & {\color[HTML]{000000} -/\underline{99.0}}    & {\color[HTML]{000000} \textbf{100}/92.9}  & {\color[HTML]{000000} 95.7/98.3} & {\color[HTML]{000000} } & {\color[HTML]{000000} 88.1/\textbf{99.1}} & {\color[HTML]{000000} 92.0/97.5}  & {\color[HTML]{000000} -/98.1}    & {\color[HTML]{000000} \textbf{100}/\textbf{99.1}}  & {\color[HTML]{000000} \textbf{100}/98.9}                           \\
{\color[HTML]{000000} Metal nut}                  & {\color[HTML]{000000} 76.1/91.8} & {\color[HTML]{000000} 88.5/92.5} & {\color[HTML]{000000} -/93.0}    & {\color[HTML]{000000} 98.7/96.3} & {\color[HTML]{000000} 96.9/93.3} & {\color[HTML]{000000} } & {\color[HTML]{000000} 71.0/\textbf{98.1}} & {\color[HTML]{000000} 94.0/\underline{98.0}}  & {\color[HTML]{000000} -/95.9}    & {\color[HTML]{000000} \textbf{100}/97.2}  & {\color[HTML]{000000} \underline{99.1}/96.8}                          \\
{\color[HTML]{000000} Pill}                       & {\color[HTML]{000000} 83.3/89.9} & {\color[HTML]{000000} 83.8/95.7} & {\color[HTML]{000000} -/97.0}    & {\color[HTML]{000000} \textbf{98.9}/97.6} & {\color[HTML]{000000} 90.2/\underline{98.3}} & {\color[HTML]{000000} } & {\color[HTML]{000000} 80.1/96.5} & {\color[HTML]{000000} 86.1/95.1}  & {\color[HTML]{000000} -/97.3}    & {\color[HTML]{000000} 96.5/97.2} & {\color[HTML]{000000} \underline{96.8}/\textbf{98.4}}                          \\
{\color[HTML]{000000} Screw}                      & {\color[HTML]{000000} 83.4/97.6} & {\color[HTML]{000000} 84.5/98.8} & {\color[HTML]{000000} -/99.0}    & {\color[HTML]{000000} 93.9/97.6} & {\color[HTML]{000000} \underline{95.7}/\textbf{99.5}} & {\color[HTML]{000000} } & {\color[HTML]{000000} 66.7/98.9} & {\color[HTML]{000000} 81.3/95.7}  & {\color[HTML]{000000} -/97.4}    & {\color[HTML]{000000} 91.8/98.7} & {\color[HTML]{000000} \textbf{98.0}/\underline{99.3}}                          \\
{\color[HTML]{000000} Toothbrush}                 & {\color[HTML]{000000} \underline{96.9}/98.1} & {\color[HTML]{000000} \textbf{100}/\underline{98.9}}  & {\color[HTML]{000000} -/98.1}    & {\color[HTML]{000000} \textbf{100}/98.1}  & {\color[HTML]{000000} \textbf{100}/\underline{98.9}}  & {\color[HTML]{000000} } & {\color[HTML]{000000} 88.9/97.9} & {\color[HTML]{000000} \textbf{100}/98.1}   & {\color[HTML]{000000} -/97.2}    & {\color[HTML]{000000} 88.6/98.6} & {\color[HTML]{000000} \textbf{100}/\textbf{99.1}}                           \\
{\color[HTML]{000000} Transistor}                 & {\color[HTML]{000000} 87.5/92.7} & {\color[HTML]{000000} 90.9/87.7} & {\color[HTML]{000000} -/80.0}    & {\color[HTML]{000000} 93.1/90.9} & {\color[HTML]{000000} 95.8/96.1} & {\color[HTML]{000000} } & {\color[HTML]{000000} 90.3/94.1} & {\color[HTML]{000000} 91.5/\underline{97.0}}  & {\color[HTML]{000000} -/90.6}    & {\color[HTML]{000000} \underline{97.8}/87.8} & {\color[HTML]{000000} \textbf{99.8}/\textbf{97.7}}                          \\
{\color[HTML]{000000} Zipper}                     & {\color[HTML]{000000} 87.5/97.8} & {\color[HTML]{000000} 98.1/97.8} & {\color[HTML]{000000} -/96.0}    & {\color[HTML]{000000} \textbf{100}/\underline{98.8}}  & {\color[HTML]{000000} \underline{99.4}/\textbf{99.2}} & {\color[HTML]{000000} } & {\color[HTML]{000000} 96.6/96.5} & {\color[HTML]{000000} 97.9/95.1}  & {\color[HTML]{000000} -/98.2}    & {\color[HTML]{000000} 97.4/98.2} & {\color[HTML]{000000} 96.0/\underline{98.8}}                          \\ \hline
{\color[HTML]{000000} Average}                    & {\color[HTML]{000000} 90.9/94.0} & {\color[HTML]{000000} 91.7/94.2} & {\color[HTML]{000000} 93.8/95.5} & {\color[HTML]{000000} \underline{98.0}/\underline{97.3}} & {\color[HTML]{000000} 95.0/96.6} & {\color[HTML]{000000} } & {\color[HTML]{000000} 85.5/96.5} & {\color[HTML]{000000} 92.1/95.7}  & {\color[HTML]{000000} 93.1/96.9} & {\color[HTML]{000000} 97.5/\underline{97.3}} & {\color[HTML]{000000} \textbf{98.7}/\textbf{98.3}}                          \\ \Xhline{1.5pt}
\end{tabular}}
\begin{tablenotes}
\footnotesize  
\item[1] {The highest image/pixel level AUROC values are in \textbf{bold}, and the second highest values are \underline{underlined}.}
\end{tablenotes}
\end{threeparttable}

\end{table*}

\begin{figure*}
    \centering
    \includegraphics[width=\textwidth]{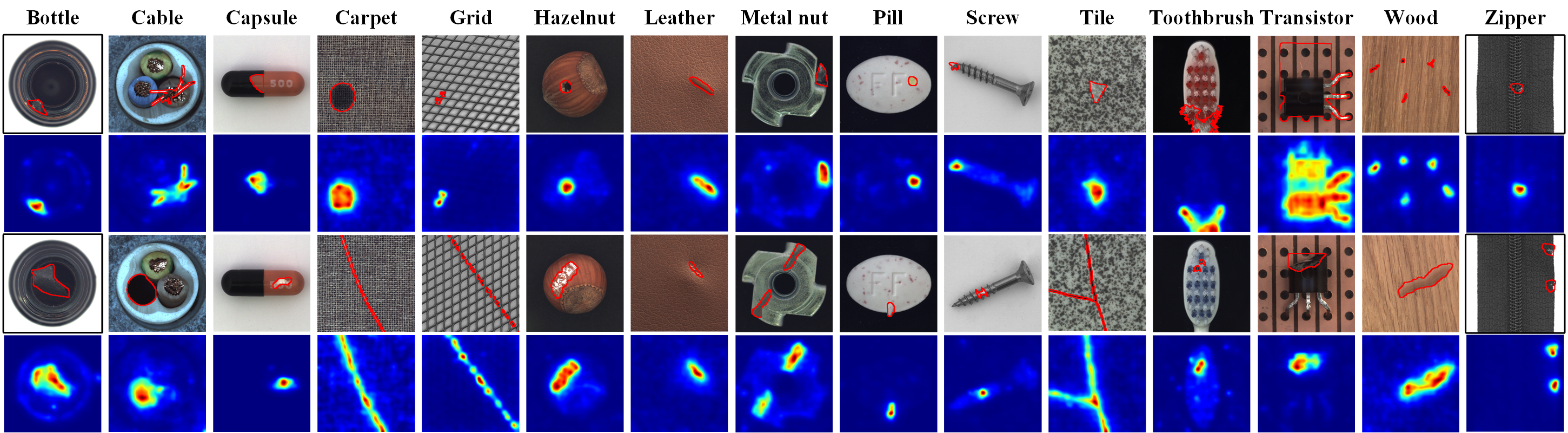}
    \caption{Segmentation results of our proposed TFA-Net on the MVTec AD dataset.}
    \label{fig:mvtec}
\end{figure*}

\section{Experiments}
In this section, we validate the overall performance of our proposed TFA-Net on the MVTec AD \cite{MVTEC} and MVTec LOCO AD \cite{mvtecloco} datasets, and conduct ablation experiments on key parameters and main modules in TFA-Net.
\subsection{Experimental Setup}
\subsubsection{Dataset Configuration}
In our experiments, we employed two datasets: MVTec AD and MVTec LOCO AD. The former is widely used for unsupervised anomaly detection and localization and comprises 15 categories, with 3629 normal samples for training and 498 normal samples and 1982 abnormal samples for testing. The latter was recently proposed for detecting logical defects and comprises 5 categories, with 1772 normal samples for training and 575 normal, 432 structural abnormal, and 561 logical abnormal samples for testing.
\subsubsection{Implement Details}
The default feature extractor in TFA-Net is the Wide-Resnet50 \cite{wideresnet} pre-trained on ImageNet \cite{ImageNet}. {The 1st to 4th layers of the pre-trained CNN are used in the feature extractor. The multi-scale feature map has a resolution of $64\times 64$, with a feature dimension of 1856.} The default patch size $K$ is 4. TFAM consists of 12 transformer blocks with hidden dimension of 768 and 12 attention heads, while FDRM contains 8 transformer blocks with hidden dimension of 512 and 16 attention heads. {The implementation of the transformer block is identical to that in ViT \cite{Vit}.} The complete TFA-Net is trained for 400 epochs using an AdamW optimizer with a learning rate of 0.001 and a batch size of 4. Each image is resized to 256 $\times$ 256 and normalized using the mean and standard deviation of the ImageNet dataset. All experiments are conducted on a computer equipped with a Intel(R) Core(TM) i5-6500 CPU running at 3.20 GHz and an NVIDIA GeForce GTX 3060 GPU with a memory size of 12 GB.
\subsubsection{Evaluation Metrics}
We utilize the widely-used the area under the receiver operating characteristic (AUROC) at image and pixel level to evaluate the detection performance of our proposed method and compared methods. Moreover, {the normalized area under the saturated per-region overlap curve (sPRO) \cite{mvtecloco} when the false postive rate is lower than 0.05 is employed for the MVTec LOCO AD dataset.} All the aforementioned metrics indicate that higher values correspond to better performance.
\subsubsection{{Comparative methods}}
{In our experiments, we compare our proposed TFA-Net with the following state-of-the-art (SOTA) methods: f-AnoGAN \cite{f-anogan}, TrustMAE \cite{TrustMAE}, RIAD \cite{RIAD}, DFR \cite{DFR}, Draem \cite{draem}, Intra \cite{Intra}, MNAD \cite{MNAD}, SPADE \cite{SPADE}, Patch SVDD \cite{PatchSVDD}, GCPF \cite{GCPF}, MBPFM \cite{MBPFM}, PatchCore \cite{PacthCore}, PaDiM \cite{PaDiM}, STPM \cite{STPM}, FastFlow \cite{fastflow}, and GCAD \cite{mvtecloco}.\\
\indent f-AnoGAN utilizes a generative adversarial network to generate more realistic reconstructed images for anomaly detection. TrustMAE employs a memory module within trust regions to reconstruct images and uses perceptual distance for anomaly detection and localization. RIAD uses a masking strategy and an image inpainting network to transform the reconstruction task into an image inpainting task. Draem proposes using natural images to synthesize artificial defect images and then employs a U-Net to predict and locate defect areas. Intra utilizes surrounding normal information to repair masked areas, resulting in defect-free reconstructed images. MNAD leverages memory-guided normality for defect detection. Patch SVDD can establish multiple normal clusters at the patch level, enabling precise anomaly localization. Both PaDiM and GCPF model normal features using multivariate Gaussian distributions, and they calculate anomaly scores by measuring the Mahalanobis distance between the test features and the Gaussian distribution. SPADE and PatchCore employ a memory bank to store normal features and determine anomalies based on the distance between test features and the most similar normal feature in the memory bank. MBPFM utilizes a multi-level bidirectional feature mapping for precise anomaly localization. STPM employs multi-scale student-teacher feature correspondence for anomaly detection. FastFlow employs probability distributions to model normal data and detects anomalies by assessing deviations from the learned distribution. GCAD enhances the model's ability to recognize logical defects by incorporating an autoencoder branch into the traditional student-teacher network.}

\subsection{Overall Comparative Experiments on MVTec AD and MVTec LOCO AD}
\subsubsection{Experiment on MVTec AD}
To verify the performance of our proposed method, we compared its detection performance with that of several SOTA methods on the MVTec AD dataset, including reconstruction-based methods such as TrustMAE\cite{TrustMAE}, RIAD \cite{RIAD}, DFR \cite{DFR}, Draem \cite{draem}, and Intra \cite{Intra}, as well as embedding-based methods such as SPADE \cite{SPADE} and Patch SVDD \cite{PatchSVDD}, GCPF \cite{GCPF}, and MBPFM \cite{MBPFM}.\\
\indent Table \ref{Table:mvtec} presents the quantitative comparison results between our proposed method and other SOTA methods. Our proposed TFA-Net achieves the best results in the average of 15 categories, with 98.7\% image-level AUROC and 98.3\% pixel-level AUROC, which outperformed the second best results by 0.7\% and 1.0\%, respectively. Notably, our method achieves an image-level AUROC of 100\% in the categories of Leather, Tile, Bottle, Hazelnut, and Toothbrush, highlighting its superior performance. Furthermore, for the challenging category of Transistor, our method achieves the best performance with an image-level AUROC of 99.8\% and a pixel-level AUROC of 97.7\%, which are 2.0\% and 0.7\% higher than the second-best results, respectively.\\
\indent The qualitative results of TFA-Net are presented in Fig. \ref{fig:mvtec}. It is evident that our method is capable of precisely localizing defects in all 15 categories. Notably, in the Cable and Transistor categories, our method is able to detect defects with object disappearance effectively, which can be attributed to the challenging and meaningful feature reconstruction task generated by our proposed TFAM, as opposed to simply replicating the input data.

\begin{figure*}
    \centering
    \includegraphics[width=\textwidth]{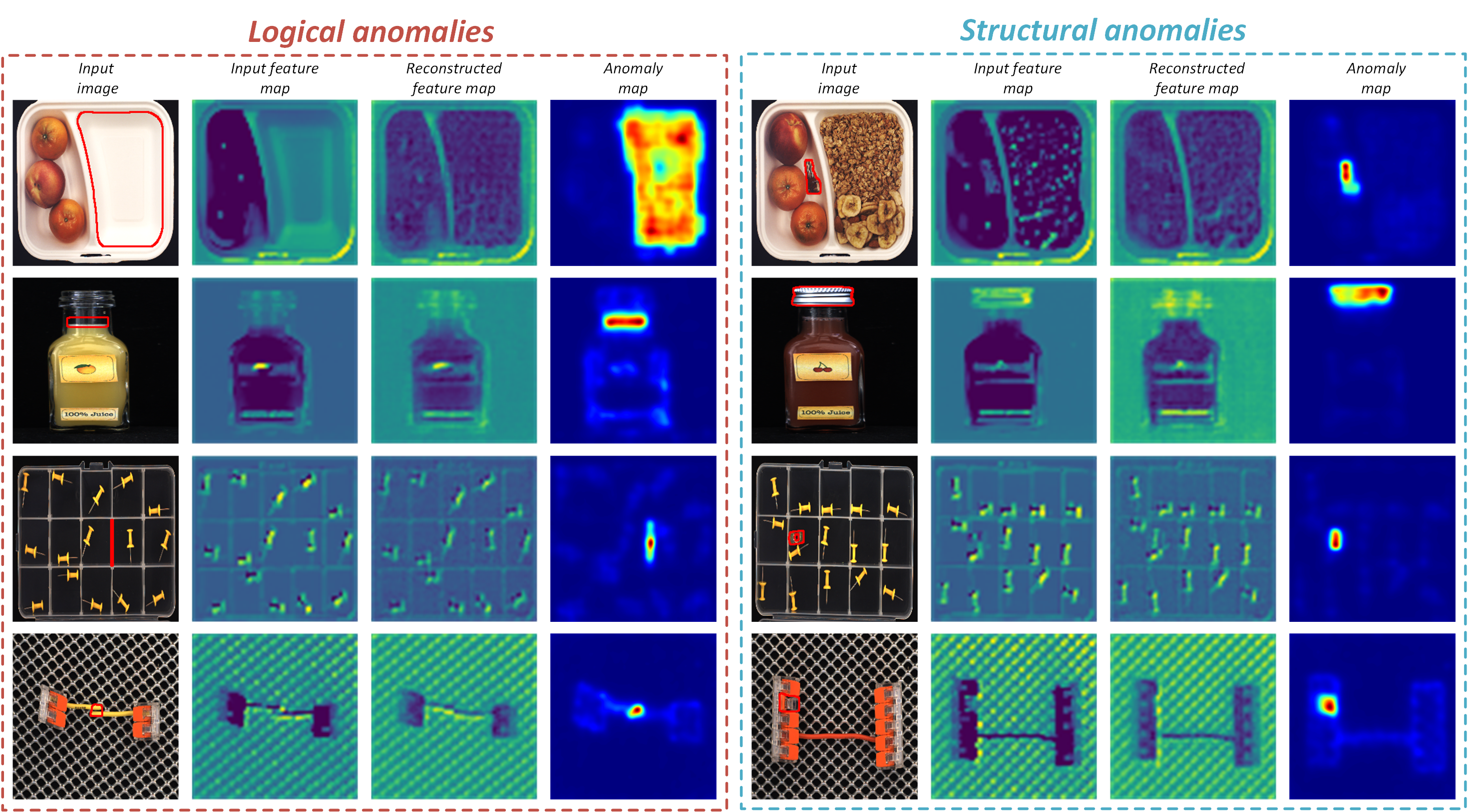}
    \caption{Reconstruction and segmentation results of our proposed TFA-Net on MVTec LOCO AD dataset.}
    \label{fig:mvtec-loco}
\end{figure*}
\begin{table}[!t]
\vspace{-1.0em}
\caption{The quantitative results of different methods on MVTec LOCO AD dataset.The experimental results of the comparative methods are sourced from \cite{mvtecloco} and \cite{IM-IAD}.}
\label{table:mvtecloco}
\setlength{\tabcolsep}{3pt}
\begin{tabular}{p{\columnwidth}}
\centering
$\includegraphics{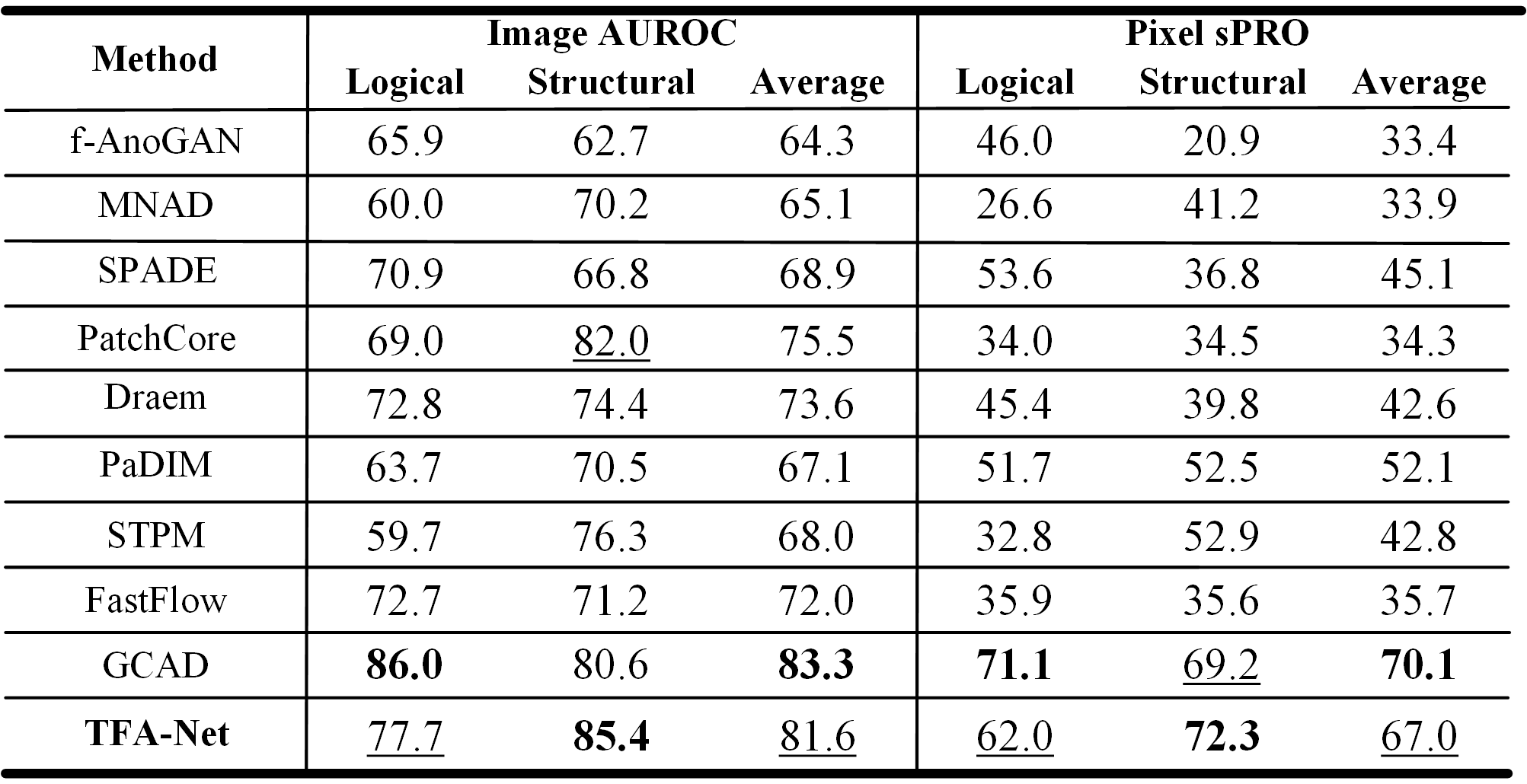}$
\end{tabular}
\vspace{-1.0em}
\end{table}

\subsubsection{Experiment on MVTec LOCO AD}
The MVTec LOCO AD dataset contains logical and structural defects, where the detection of logical defects is challenging and requires high semantic information. Therefore, to further validate the effectiveness of our proposed method, we compared our method with several SOTA methods on the MVTec LOCO AD dataset, including f-AnoGAN \cite{f-anogan}, MNAD \cite{MNAD}, SPADE \cite{SPADE}, PatchCore \cite{PacthCore}, 
Draem \cite{draem}, PaDIM \cite{PaDiM}, STPM \cite{STPM}, FastFlow\cite{fastflow}, and GCAD \cite{mvtecloco}. \\
\indent Table \ref{table:mvtecloco} presents the quantitative comparison results of the experiments. TFA-Net outperforms the other methods in detecting structural anomalies, achieving image AUROC of 85.4\% and pixel sPRO of 72.3\%. In detecting logical anomalies, TFA-Net ranks second, following GCAD. Nonetheless, it should be emphasized that GCAD is an approach specifically tailored for the MVTec LOCO AD dataset, and it only achieves an image AUROC of 93.1\% on the MVTec AD dataset, which is significantly lower than the performance of our proposed TFA-Net. \\
\indent The qualitative experimental results are presented in Fig. \ref{fig:mvtec-loco}. Our proposed method showcases its ability to effectively repair both logical and structural defects with precise localization. This performance is largely attributed to the innovative introduction of TFAM, which facilitates the consideration of global semantic information and allows the model to obtain meaningful reconstruction results, instead of merely replicating the input feature.

\subsection{Ablation Experiments on MVTec AD}
\subsubsection{The influence of feature extractor}
TFA-Net adopts a pre-trained feature extractor to extract multiple hierarchical fused features, which serve as the reconstruction target. The goal is to obtain features that are rich in semantic information, preserve spatial details, and possess the ability to discriminate defects. The discriminative power of pre-trained networks can vary. Therefore, the impact of this feature extractor is extensively examined in the following analysis. \\
\indent Table \ref{table:influence-feature-extrator} presents the impact of different feature extractors (MobileNet, VGG19, WideResnet50, and WideResnet101) on the performance and efficiency of the proposed model. Using WideResnet101 as the feature extractor yields the best detection results, as its extracted features contain rich semantic information. Notably, the detection performance obtained using WideResnet50 is only slightly inferior to the best result, while offering a higher FPS than WideResnet101 by 5.2 and significantly reducing parameter and computational complexity. Therefore, to achieve a better balance between detection performance and speed, we choose to use WideResnet50 as the feature extractor.
\begin{figure}
    \centering
    \includegraphics{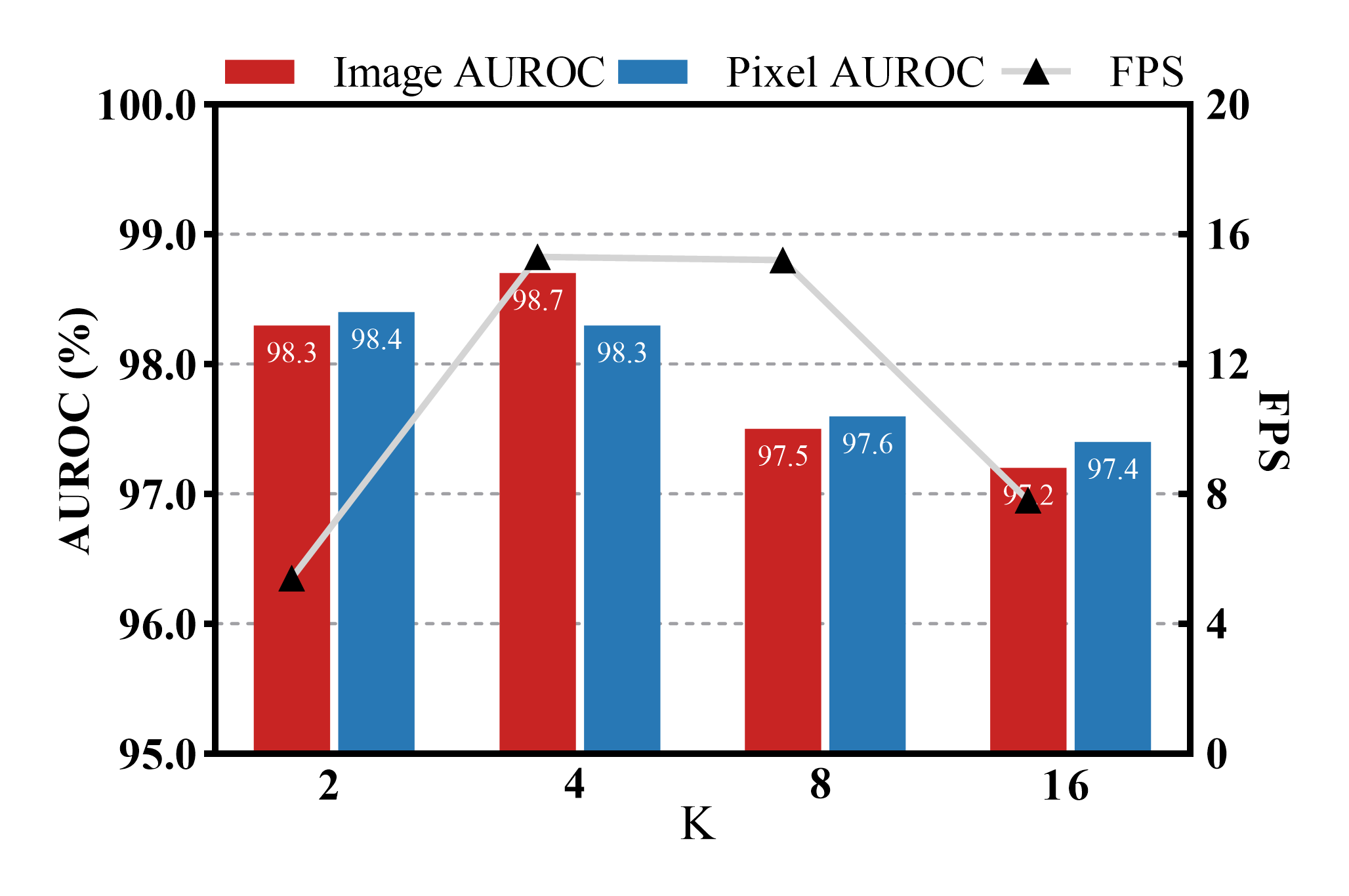}
    \caption{The impact of {patch size} $K$ on the performance and inference time of the model.}
    \label{fig:influence-K}
    \vspace{-1.0em}
\end{figure}
\subsubsection{The influence of patch size $K$}
TFA-Net partitions the multi-scale fused feature map into a series of local semantic descriptors using patch size $K$. The size of $K$ can affect the detection performance and inference speed of the model. Therefore, a detailed analysis of the size of $K$ is presented in the following.\\
\indent The influence of different sizes of $K$ (2, 4, 8, and 16) is shown in Fig. \ref{fig:influence-K}. When $K$ is small, the granularity of the local semantic descriptors is finer, allowing for better feature aggregation and more precise reconstruction results. However, this also leads to longer sequences and significantly increased computational complexity, resulting in slower inference speed of the model. When $K$ is too large, it not only leads to coarse reconstruction results but also increases the number of parameters in the linear projection head of ViT \cite{Vit}, resulting in a decrease in inference speed. To strike a balance between inference time and detection performance, we set the value of $K$ to 4.

\begin{table}[t]
\vspace{-1.0em}
\caption{The impact of feature extractors on the performance and efficiency of the model}
\label{table:influence-feature-extrator}
\begin{threeparttable}
    \setlength{\tabcolsep}{3pt}
\begin{tabular}{p{\columnwidth}}
\centering
$\includegraphics[width=100mm]{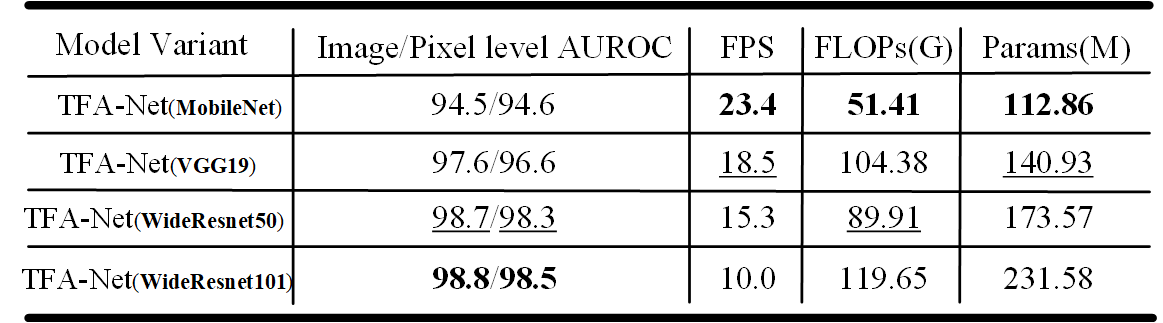}$
\end{tabular}
\begin{tablenotes}
\footnotesize  
\item[1] {The units for FLOPs and Params are $G$ and $M$, where $1G = 10^9$ and $1M = 10^6$.}
\end{tablenotes}
\vspace{-1.0em}
\end{threeparttable}
\end{table}

\subsubsection{The robustness to template images}
\label{tempate-ablate}
\begin{figure}[!t]
    \centering
    \includegraphics[width=170mm]{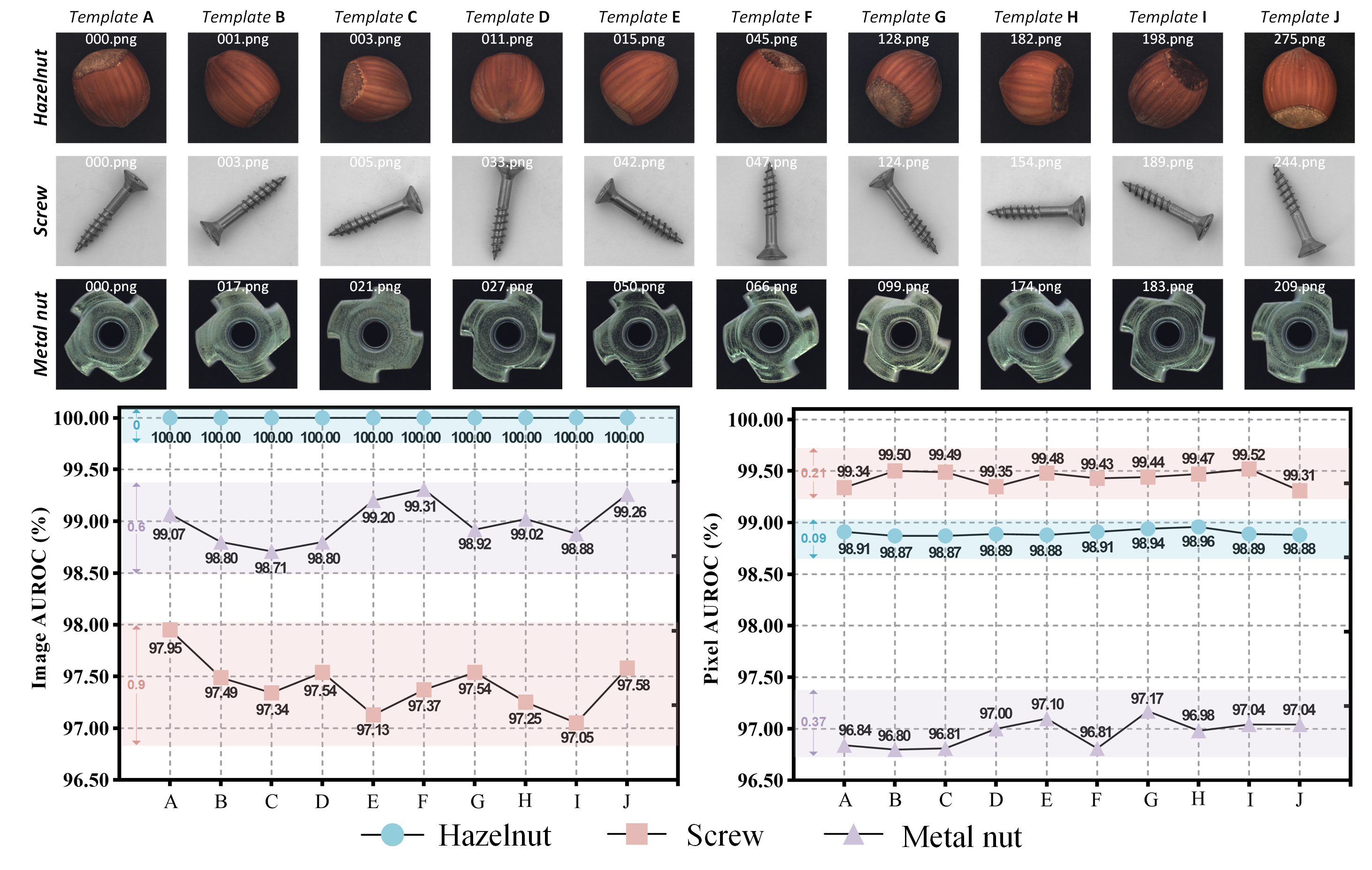}
    \caption{The impact of template image selection on the performance of the model for Hazelnut, Screw, and Metal nut in the MVTec AD dataset.}
    \label{fig:template-ablation}
\end{figure}

TFA-Net demonstrates robustness in template image selection, indicating that different template images do not significantly affect the performance of TFA-Net. To validate the robustness of model performance in template image selection, {we conducted ablation experiments using three datasets with pose diversity from the MVTec AD dataset: Hazelnut, Screw, and Metal nut. For each dataset, we selected 10 normal images with different poses as template images and performed end-to-end training using the corresponding templates.}\\
\indent {The selected template images quantitative experimental results are presented above and below, respectively, in Fig. \ref{fig:template-ablation}. Notably, even when opting for entirely different normal images as templates, TFA-Net maintains excellent performance on the Hazelnut, Screw, and Metal Nut datasets. In more detail, on the Hazelnut dataset, the model exhibits a maximum fluctuation of only 0/0.09\% in Image/Pixel level AUROC. On the Screw dataset, the model shows a maximum fluctuation of merely 0.9/0.21\% in Image/Pixel level AUROC. Similarly, on the Metal nut dataset, the model demonstrates a maximum fluctuation of only 0.6/0.37\% in Image/Pixel level AUROC. The fluctuation in the model's performance on the Screw and Metal Nut datasets is noticeably greater than its fluctuation on the Hazelnut dataset. This discrepancy is attributed to the fact that, in comparison to the Hazelnut dataset, the Screw and Metal Nut datasets exhibit significantly greater diversity, resulting in larger fluctuations in the model's performance. In general, in the three data sets with pose diversity, the variations in model performance remain below 1\%. We consider such minor fluctuations to be within an acceptable range.}  These results provide substantial evidence for the robustness of TFA-Net in template image selection.

\begin{figure}[t]
    \centering
    \includegraphics{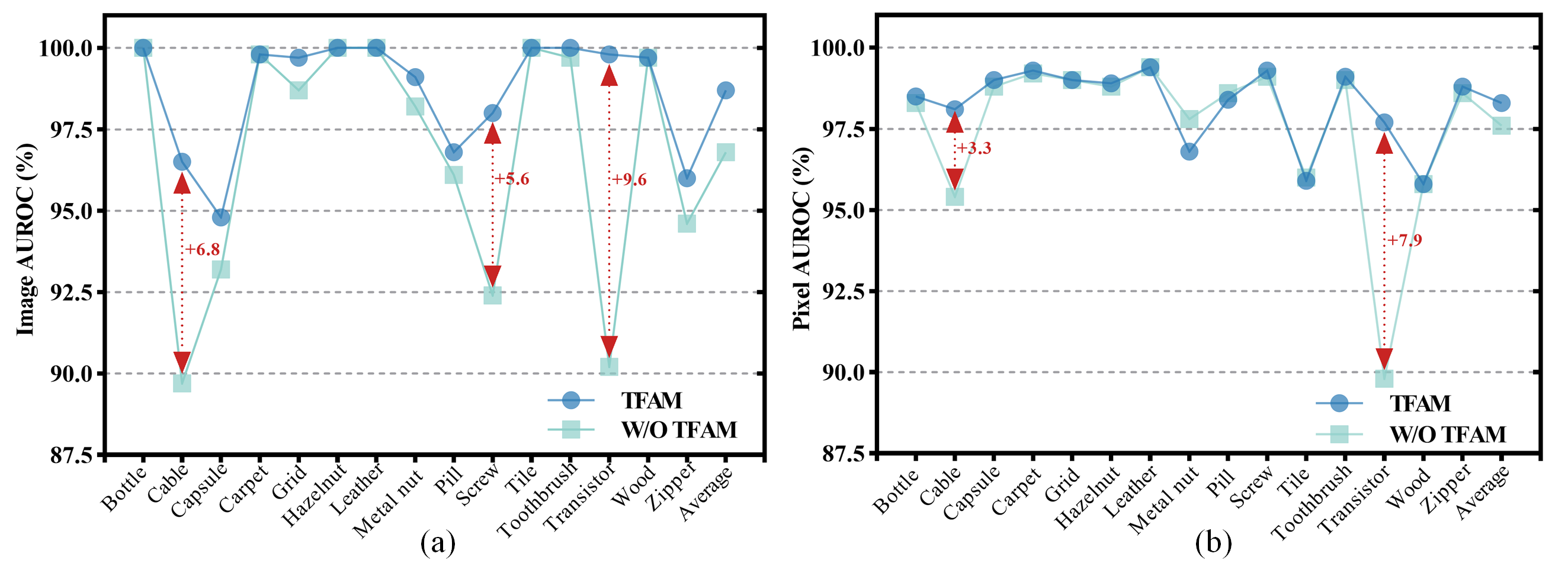}
    \caption{The influence of TFAM on the performance of the model. (a) Image-level AUROC results. (b) Pixel-level AUROC results.}
    \label{fig:influence-TFAM}
\end{figure}

\begin{figure}[t]
    \centering
    \includegraphics{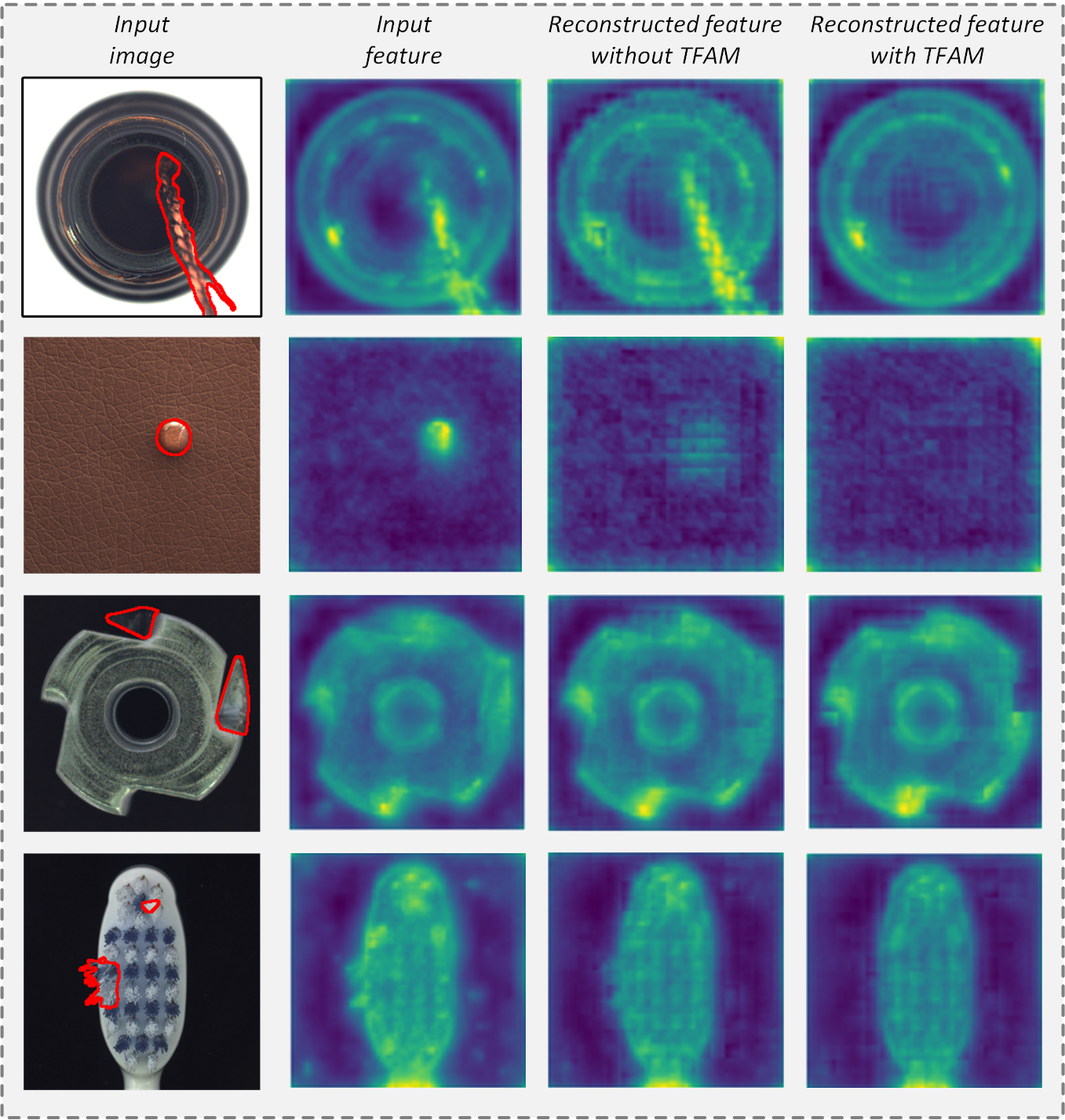}
    \caption{Some examples of the influence of template-based feature aggregation mechanism (TFAM).}
    \label{fig:TFAM-example}
\end{figure}
\subsubsection{The influence of TFAM}
TFAM is designed to aggregate normal information from input features into template features and filter out abnormal information, transforming the simple feature reconstruction task into a challenging feature aggregation task and producing meaningful reconstruction results. To validate the effectiveness of TFAM, we compare it with the vanilla ViT (without TFAM).\\
\indent Fig. \ref{fig:influence-TFAM} presents the quantitative comparison experimental results. The results show that TFAM outperforms the model without TFAM in terms of average image/pixel AUROC across 15 categories on MVTec AD dataset. It is worth noting that in image-level detection (as shown in Fig. \ref{fig:influence-TFAM}(a)), TFAM significantly improved the performance by 6.8\%, 5.6\%, and 9.6\% for Cable, Screw, and Transistor categories, respectively, compared to the model without TFAM. In pixel-level detection (as shown in Fig. \ref{fig:influence-TFAM}(b)), TFAM also showed significant improvements by 3.3\% and 7.9\% for Cable and Transistor categories, respectively, compared to the model without TFAM. {Overall, model with TFAM demonstrates a significant improvement in anomaly detection and localization performance on datasets such as Transistor and Cable compared to model without TFAM. This is attributed to TFAM enabling the model to learn more semantically rich information, allowing for the detection of global defects in datasets like Transistor and Cable, such as object omissions, which require a high level of semantic information for accurate detection. However, model with TFAM exhibits a slight decrease in detection performance on datasets like Metal Nut and Pill compared to model without TFAM. This is because the introduction of TFAM may lead to the loss of some details in the reconstructed feature map, resulting in the emergence of noise in the anomaly map and consequently causing a reduction in detection performance.}\\
\indent Fig. \ref{fig:global-defect} and Fig. \ref{fig:TFAM-example} illustrate examples showcasing the impact of TFAM. The model without TFAM achieves impeccable reconstruction of defect features, while the model integrated with TFAM effectively filters out defect features, yielding reconstructed features that are devoid of anomalies. Consequently, this augmentation significantly bolsters the detection accuracy of the model.

\begin{table}[t]
\caption{The impact of dual-mode anomaly segmentation on the performance of the model}
\label{table:dual-mode}
\setlength{\tabcolsep}{3pt}
\begin{tabular}{p{\columnwidth}}
\centering
$\includegraphics[width=88mm]{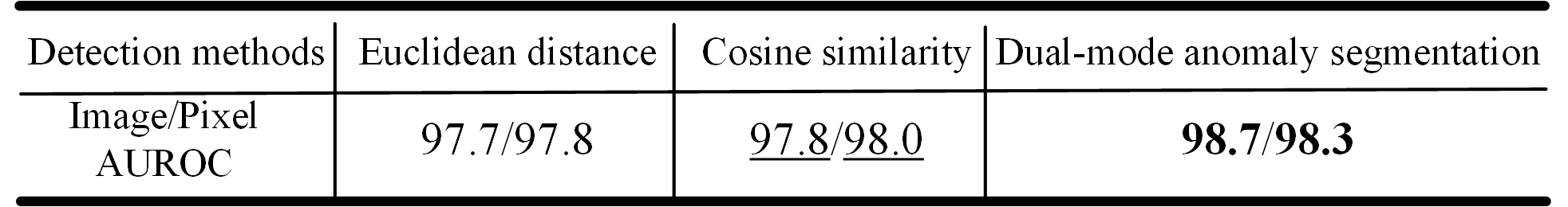}$
\end{tabular}
\end{table}
\begin{figure}
    \centering
    \includegraphics{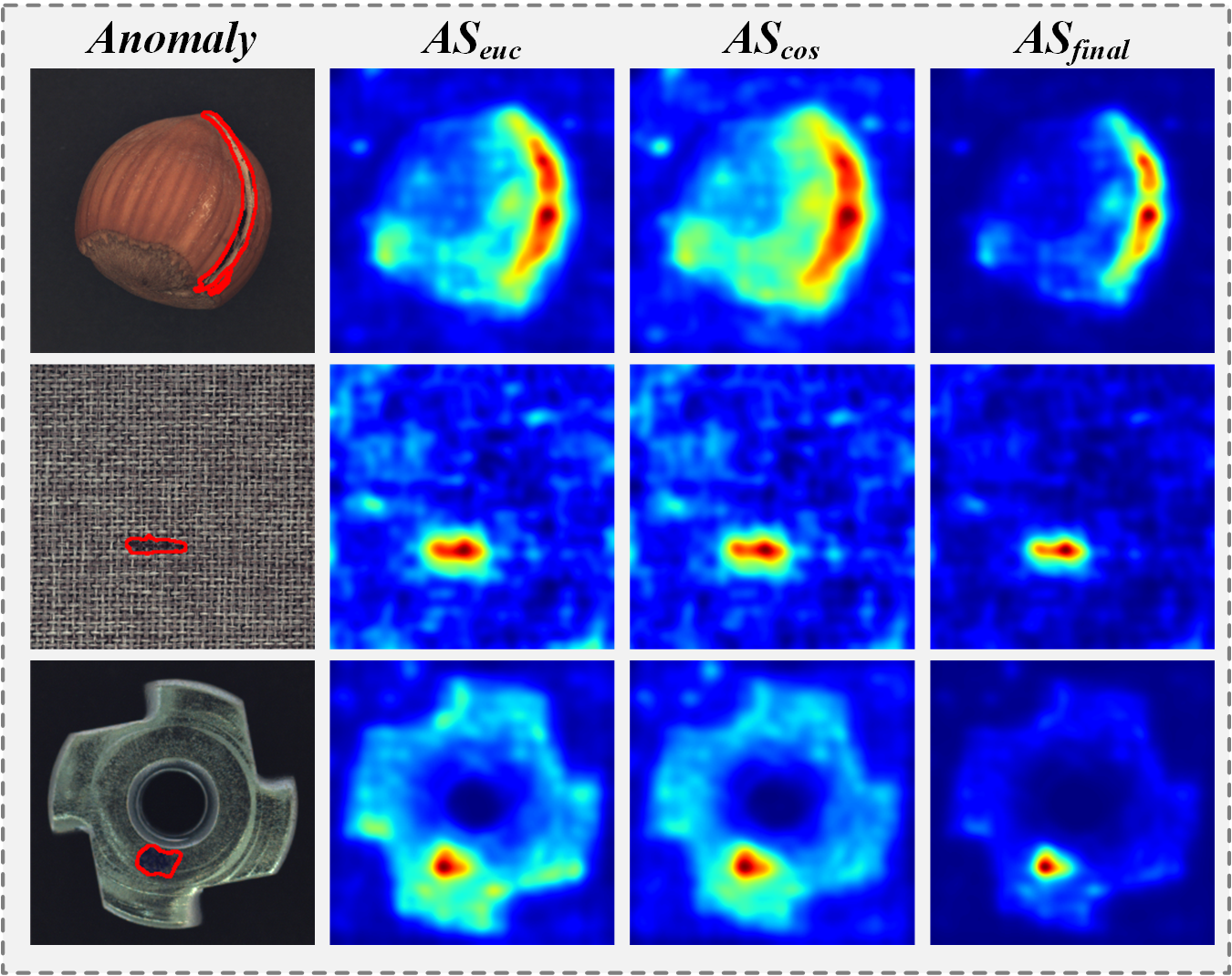}
    \caption{Some examples of the effect of the dual-mode anomaly segmentation method.}
    \label{fig:dual_mode_examples}
\end{figure}
\subsubsection{The influence of dual-mode anomaly segmentation}
The dual-mode anomaly segmentation method is crucial for accurately localizing anomalous regions based on input and reconstructed features. Therefore, in the following, we analyze its impact in detail.\\
\indent Table \ref{table:dual-mode} presents the influence of the dual-mode anomaly segmentation on the model's detection performance. Compared to using solely Euclidean distance or cosine similarity, the dual-mode anomaly segmentation method improves the image/pixel AUROC by 1.0\%/0.5\% and 0.9\%/0.3\%, respectively. Fig. \ref{fig:dual_mode_examples} showcases qualitative comparison results, demonstrating that the dual-mode anomaly segmentation method accurately localizes defects and effectively suppresses noise.
\begin{figure}[!t]
    \centering
    \includegraphics{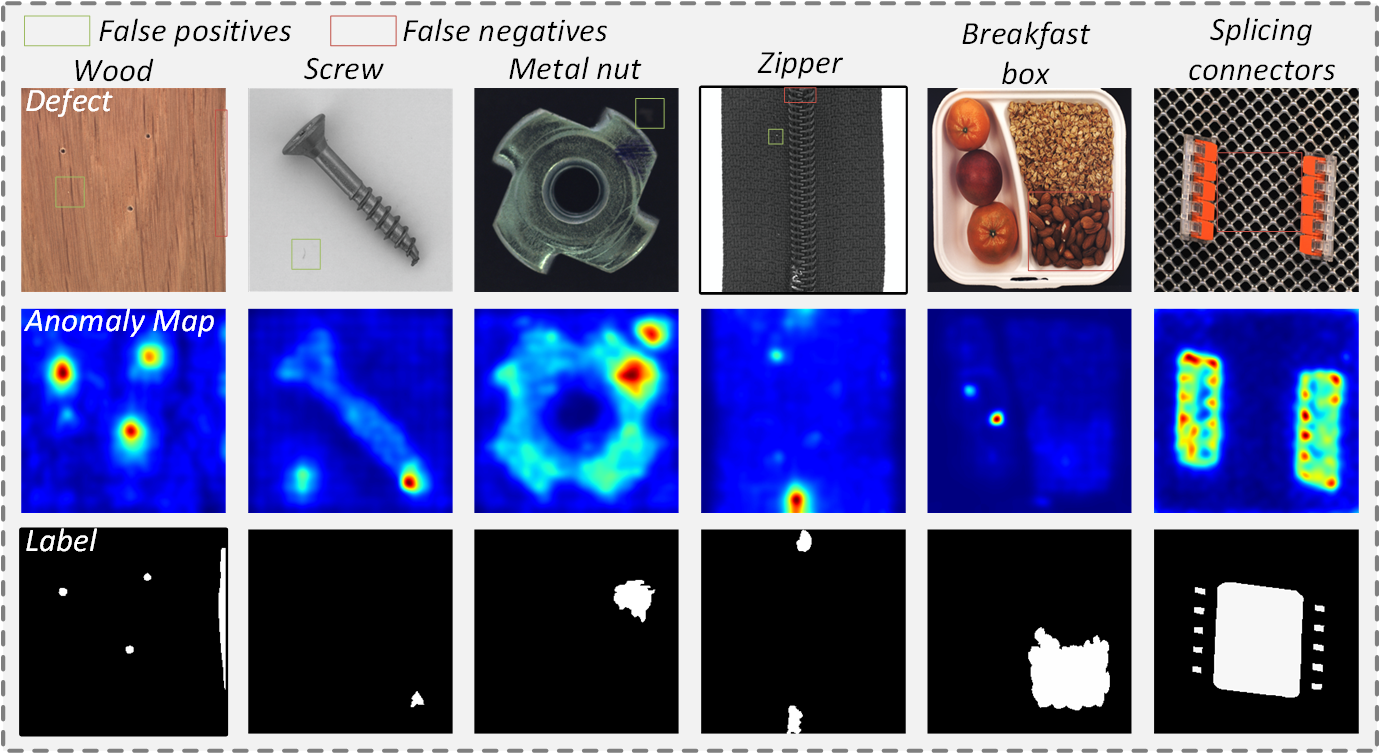}
    \caption{Some failure cases of the TFA-Net method.}
    \label{fig:failure}
\end{figure}
\subsection{Analysis of Failure Cases}
The above-mentioned ablation experiments and comparative analyses convincingly establish the superiority of our proposed method compared to existing approaches. Nevertheless, it is crucial to recognize that our method is not without limitations.\\
\indent As illustrated in Fig. \ref{fig:failure}, our method may encounter instances of both false positives and false negatives during the detection process. Specifically, when interference is present in the background, our method might erroneously identify it as a defect, leading to false positive occurrences, as observed in the Wood, Screw, and Metal nut datasets. Moreover, when the contrast between the defects and the background is low, the occurrence of false negatives becomes evident, as seen in the Wood and Zipper datasets. This phenomenon can be attributed to the utilization of a pre-trained network on the ImageNet \cite{ImageNet} dataset as a feature extractor in our model, which may lack optimal discriminative capabilities for specific textures and objects present in these datasets. Furthermore, our method exhibits limitations in detecting complex logical defects, as evidenced in the Breakfast box and Splicing connectors datasets. This emphasizes the importance of devising more meaningful training tasks for anomaly detection methods to enable the model to learn deeper semantic information.\\
\indent The above-mentioned observations underscore the necessity for further research, highlighting the importance of improving the model's discriminative capacity and refining the learning tasks to enhance its robustness and efficacy.
\section{Conclusion}
This paper presents a novel feature reconstruction network, named TFA-Net, that aims to achieve accurate anomaly detection and segmentation. The key innovation of this network lies in the introduction of TFAM, which utilizes a fixed normal image as a template and applies the multi-head self-attention mechanism of ViT to aggregate normal information from the input features onto the template features. This process effectively filters out abnormal information and produces meaningful reconstruction results that go beyond mere duplication of the input data. Subsequently, a dual-mode anomaly segmentation approach is employed to compute the discrepancies between the input features and the reconstructed features. Overall, TFA-Net offers a promising solution for the detection and segmentation of anomalies, and our results demonstrate its effectiveness and potential for future applications. In the future, we will investigate further enhancements of the performance of this network on logical anomalies.

\section*{CRediT authorship contribution statement}
\textbf{Wei Luo}: Conceptualization, Methodology, Formal analysis, Investigation, Writing–original draft, Writing–review \& editing, Visualization. 

\textbf{Haiming Yao}: Methodology, Writing–review \& editing, Validation, Supervision. 

\textbf{Wenyong Yu}: Methodology, Writing-review \& editing.

\section*{Declaration of competing interest}
The authors declare that they have no known competing financial interests or personal relationships that could have appeared to
influence the work reported in this paper.
\section*{Data availability}
The data is available online.
\section*{Acknowledgments}
This study was financially supported by the National Natural Science Foundation of China (Grant No. 52375494).
\bibliographystyle{unsrt}

\bibliography{cas-refs}

\end{document}